\documentclass[numbered]{trbunofficial}
\usepackage{pdflscape}
\usepackage{graphicx}
\usepackage{subcaption}
\usepackage{float}
\usepackage{booktabs}
\usepackage{multirow}
\usepackage{array}
\usepackage[table,xcdraw]{xcolor}
\usepackage{pdfpages}
\usepackage{gensymb}
\usepackage[hidelinks]{hyperref}

\begin{document}
\nolinenumbers
% ---------- Title Page ----------
% Place Title Page AFTER \begin{document} and BEFORE \maketitle 
% to include the front matter in word count

% Paper title
\title{Controlled Experiments on Lane Changing by Transitional Autonomous Vehicle: Dataset and Behavioral Insights}

% Add author(s) with: \TRBauthor[*]{Name}{Affiliation}{Email}[Address][ORCID]

\TRBauthor{Abhinav Sharma}
{Department of Civil, Construction, and Environmental Engineering, North Carolina State University}
{asharm63@ncsu.edu}
[Raleigh, NC 27695, USA]

\TRBauthor{Md Abdullah Al Hasan}
{Department of Civil, Construction, and Environmental Engineering, North Carolina State University}
{malhasa2@ncsu.edu}
[Raleigh, NC 27695, USA]

\TRBauthor*{Danjue Chen}
{Department of Civil, Construction, and Environmental Engineering, North Carolina State University}
{dchen33@ncsu.edu}
[Raleigh, NC 27695, USA]

\TRBauthor{George F. List}
{Department of Civil, Construction, and Environmental Engineering, North Carolina State University}
{gflist@ncsu.edu}
[Raleigh, NC 27695, USA]

% Short author line for the running header
\AuthorHeaders{Sharma, Al Hasan, Chen, and List}

% Optional title-page disclaimer/copyright
% \TRBtitlefootnote{%
%   This \LaTeX\ template is unofficial and may not meet current TRB requirements. Please review the latest submission rules at \href{https://trb.secure-platform.com/a/page/TRBPaperReview}{TRB’s Instructions for Authors} and ensure your paper is compliant. Noncompliant papers may be rejected.  Report any issues to \url{https://github.com/chiehrosswang/TRB_LaTeX_tex}.
% }

\maketitle

% ---------- Abstract ----------
\section{Abstract}

\noindent\textbf{Objectives:} This paper presents the North Carolina Transitional Autonomous Vehicle Lane-Changing (NC-tALC) dataset and uses it to characterize mandatory lane-changing behavior of a transitional automated vehicle (tAV). The objectives are to quantify the evolution of lead--lag gaps throughout the lane-change process and to identify how potential collision risk develops and changes during the maneuver.

\hfill\break
\noindent\textbf{Methods:} A controlled field experiment consisting of 78 mandatory lane-change trials was conducted on a public roadway in Apex, North Carolina. Four instrumented vehicles created repeatable initial traffic conditions while varying the lane changer's initial position within the candidate target gap. High-resolution RTK-GNSS/INS trajectories were obtained to identify key lane-change timestamps, compute lead, lag, and lane-change gaps, and estimate vehicle interactions using time-gap- and speed-based surrogate safety measures.

\hfill\break
\noindent\textbf{Findings:} Despite substantial differences in the initial conditions, the lead and lag gaps consistently evolved toward a relatively narrow range around lane crossing, indicating a convergence behavior during the lane-change process. Potential collision risk increased as the maneuver progressed, peaked near physical lane entry, and was dominated by interactions with the target-lane leader. The analysis also showed that lane-change completion does not necessarily correspond to the disappearance of collision risk.

\hfill\break
\noindent\textbf{Novelty:} This paper provides one of the first controlled empirical characterizations of the complete mandatory lane-change process of transitional automated vehicles using repeatable public-road experiments. The NC-tALC dataset enables analysis of both behavioral evolution and safety evolution throughout the maneuver rather than only at the gap-acceptance instant.

\hfill\break
\noindent\textbf{Practical Applications:} The dataset and findings provide empirical benchmarks for evaluating automated lane-changing behavior, calibrating behavioral models, and validating simulation and safety assessment methods for transitional automated vehicles operating in mandatory lane-change scenarios.

\medskip
\noindent\textit{Keywords:} transitional autonomous vehicle; lane changing;  controlled experiment; behavior; safety.

\newpage

% ---------- Introduction ----------
% \section{Introduction}\label{sec:intro}
% The Transportation Research Board (TRB) currently requires submissions of full papers to be considered for presentation at the TRB Annual Meeting \cite{trbwebsite}.  The Instructions For Authors website (\url{https://trb.secure-platform.com/a/page/TRBPaperReview}) outlines specific requirements for submissions. Initial submissions are PDFs, while accepted papers for the \textit{Transportation Research Record} may require Microsoft Office formats. Manuscripts must be line-numbered; captions are bold with TRB-specific punctuation; in-text citations now use author--date format, and the reference list is alphabetical rather than numbered. See the author information online at \url{https://trb.secure-platform.com/a/page/TRBPaperReview}.

% We assume basic familiarity with \LaTeX\ and \verb|bibtex|. As literate programming becomes more common, the template may evolve to support additional workflows.

\newcommand{\figTestSite}{\begin{figure}[t]
  \centering
  \begin{subfigure}[t]{0.48\linewidth}
    \centering
    \includegraphics[width=\linewidth, height = 6 cm]{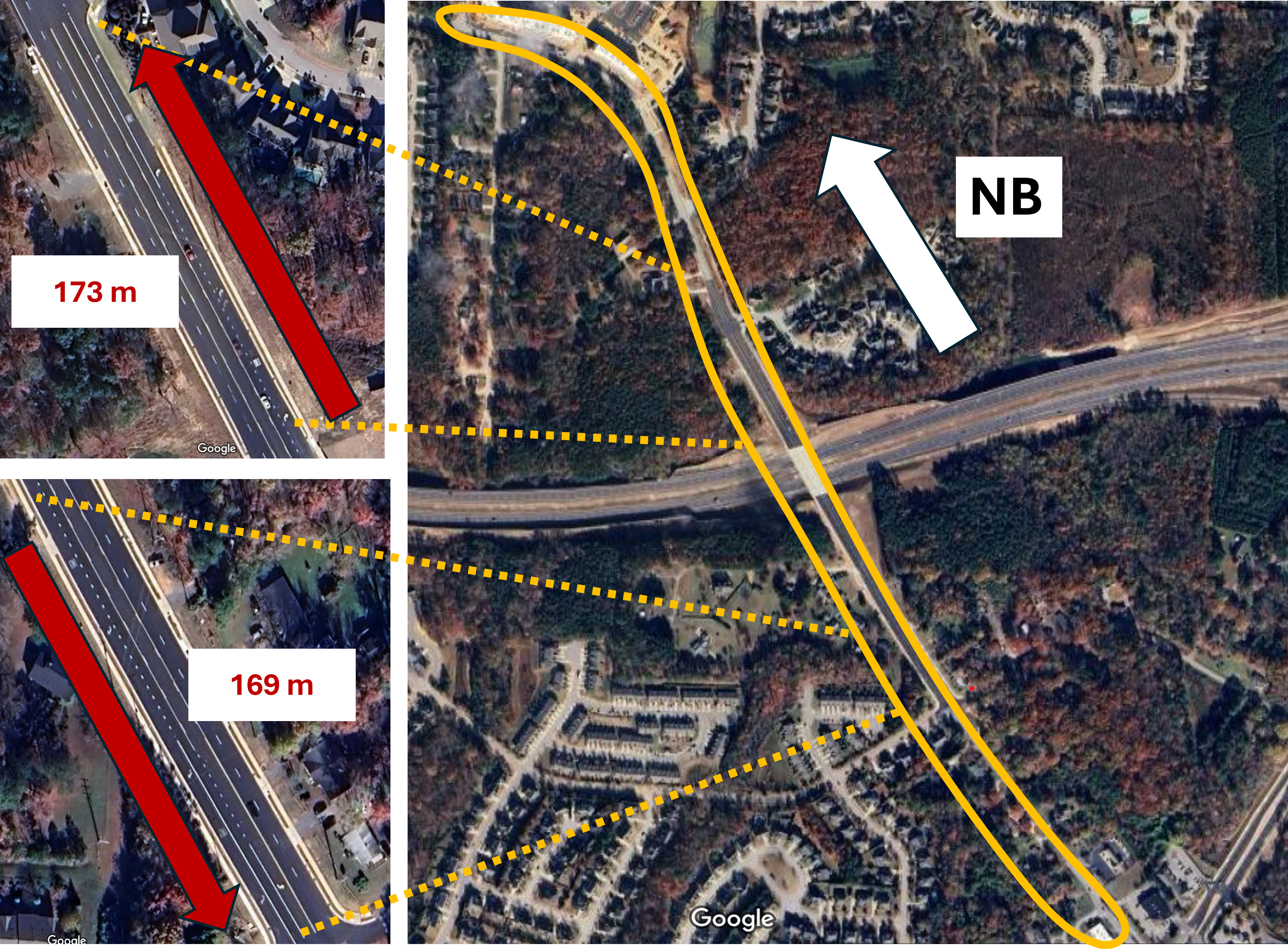}
    \caption{Test segment in NB and SB Direction; Google Maps (35.674718, -78.804330)}
    \label{fig:left}
  \end{subfigure}\hfill
  \begin{subfigure}[t]{0.48\linewidth}
    \centering
    \includegraphics[width=\linewidth, height = 6 cm]{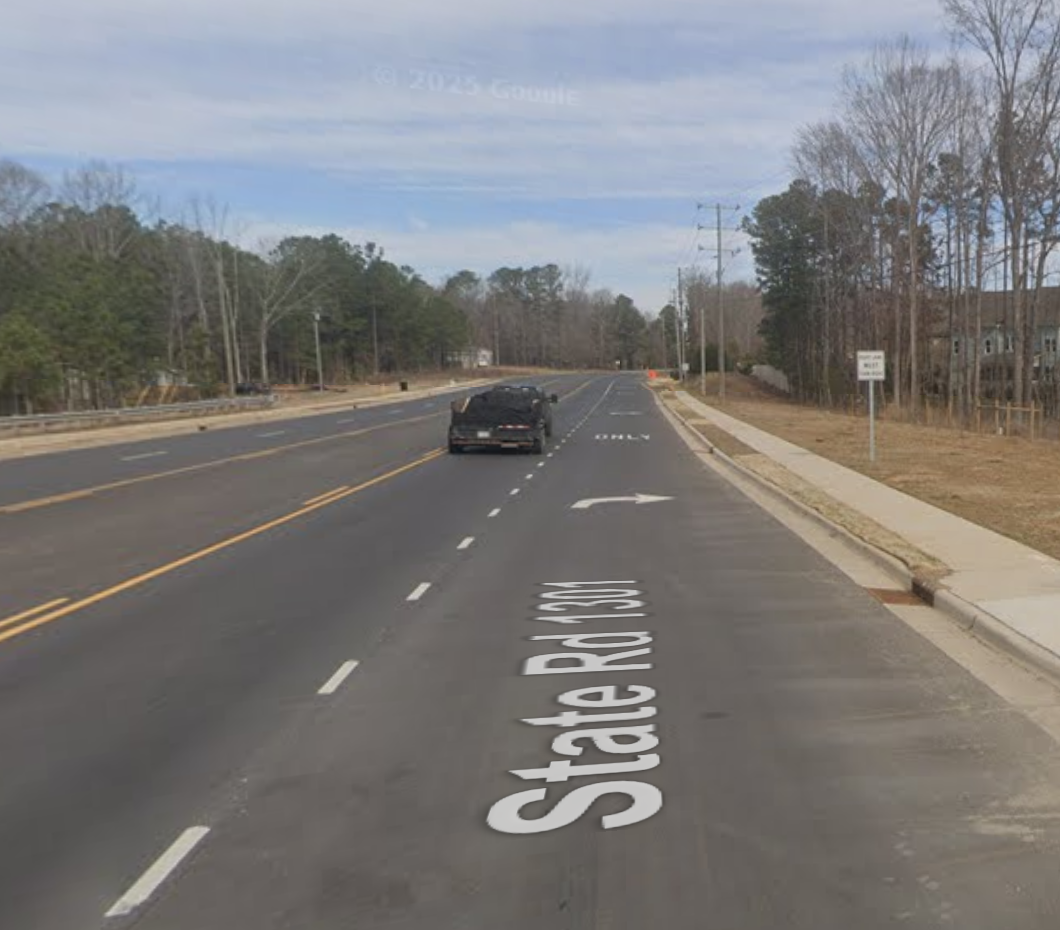}
    \caption{Right-turn lane in NB Direction}
    \label{fig:right}
  \end{subfigure}
  \caption{Experiment site in Apex, North Carolina, USA}
  \label{fig:Test Site}
\end{figure}}

\newcommand{\SiteLayout}{
\begin{figure}[t]
    \centering
    \includegraphics[width=\linewidth]{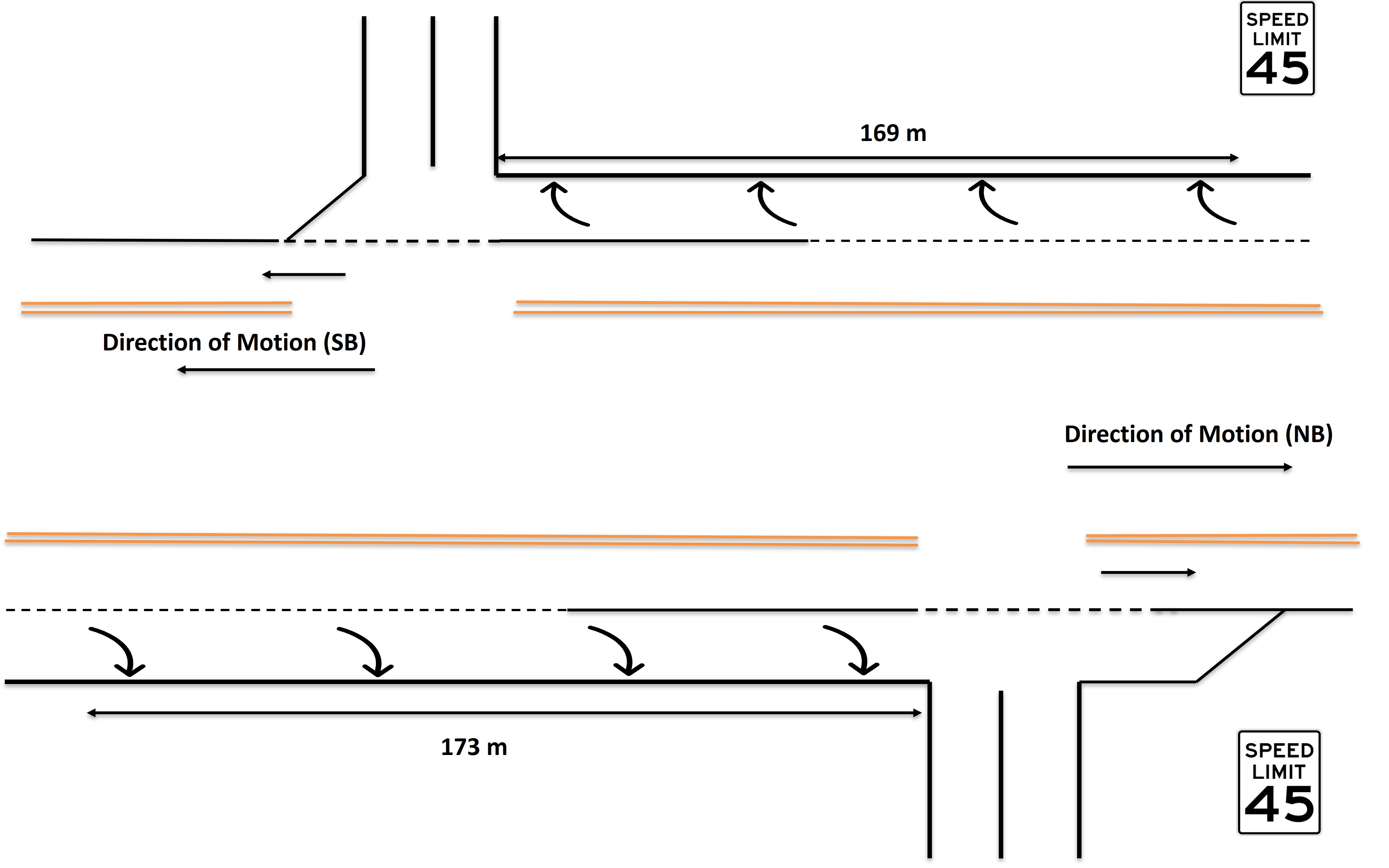}
    \caption{Geometric layout}
    \label{fig:Site Layout}
\end{figure}}

\newcommand{\figVehicleConfig}{
\begin{figure}[t]
    \centering
    \includegraphics[width=\linewidth,height=6cm]{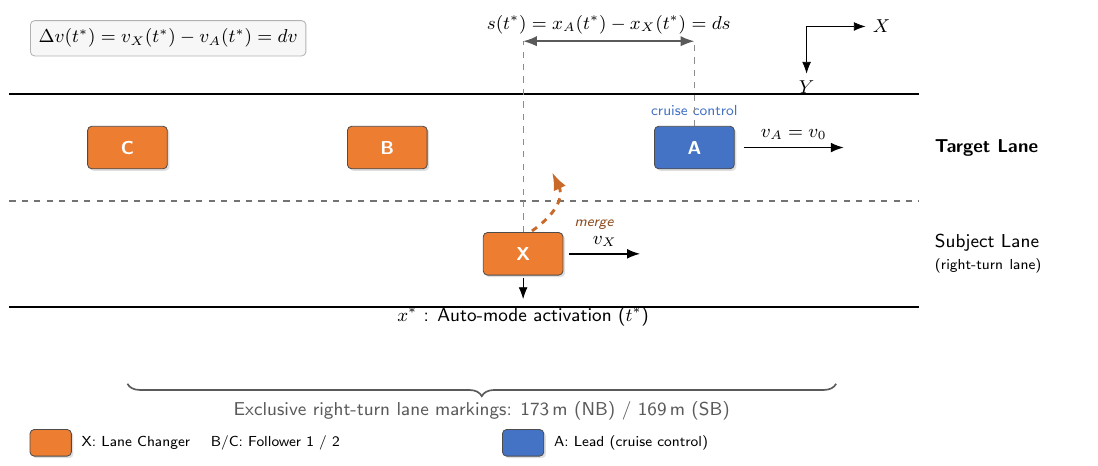}
    \caption{Vehicle configuration and definition of the controlled initial conditions. At Auto-mode activation (time $t^{*}$, location $x^{*}$) the relative spacing $ds = x_l(t^{*}) - x_{lc}(t^{*})$ and relative speed $dv = v_{lc}(t^{*}) - v_l(t^{*})$ of the lane-changing (LC) vehicle with respect to the target lead are recorded; B and C are the target-lane followers. The exclusive right-turn lane provides 173~m (NB) / 169~m (SB) of mandatory-merge distance.}
    \label{fig:Vehicle Configuration}
\end{figure}}

\newcommand{\ReferenceMaps}{
\begin{figure}[t]
    \centering
    \includegraphics[width=\linewidth]{Figures/ReferenceMaps.jpg}
    \caption{Lane Centerlines NB and SB direction with intermittent exclusive right-turn segment}
    \label{fig:Reference Maps}
\end{figure}}

\newcommand{\NGSIMtotalgap}{
\begin{figure}[t]
    \centering
    \includegraphics[width=\linewidth]{Figures/NGSIMLeadgapvstotalgap.jpg}
    \caption{NGSIM Lead Gap vs Target Gap; (1) US101 Lane 6 to Lane 5 mandatory lane change cases (2) Only constrained target Gap (< 5 seconds) (3) Sample Size = 28}
    \label{fig:NGSIMtotalgap}
\end{figure}}

\newcommand{\IndependentVariablesSelection}{
\begin{figure}[t]
  \centering
  \includegraphics[width=\linewidth]{Figures/fig_independent_vars.pdf}
  \caption{Selection of the independent variables. The target gap ($H \approx 1.5$~s $\approx 30$~m at 40~mph) is partitioned into an \textit{easy} (gap-acceptable) and a \textit{difficult} (gap-rejected) merging zone. The NGSIM-derived boundary at a 0.9~s lead gap (dashed baseline) is hypothesized to shift with relative speed: a speed advantage ($dv>0$) enlarges the easy zone toward larger gaps, whereas a speed deficit ($dv<0$) enlarges the difficult zone toward smaller gaps. The discrete spacing categories (DS0–DS3) used for sampling are aligned to the same lead-gap axis (cf. Table~\ref{tab:sample_distribution_combined}).}
  \label{fig:independent_variables_selection}
\end{figure}}

\newcommand{\OverlayTrajectories}{
\begin{figure}[t]
    \centering
    \includegraphics[width=\linewidth]{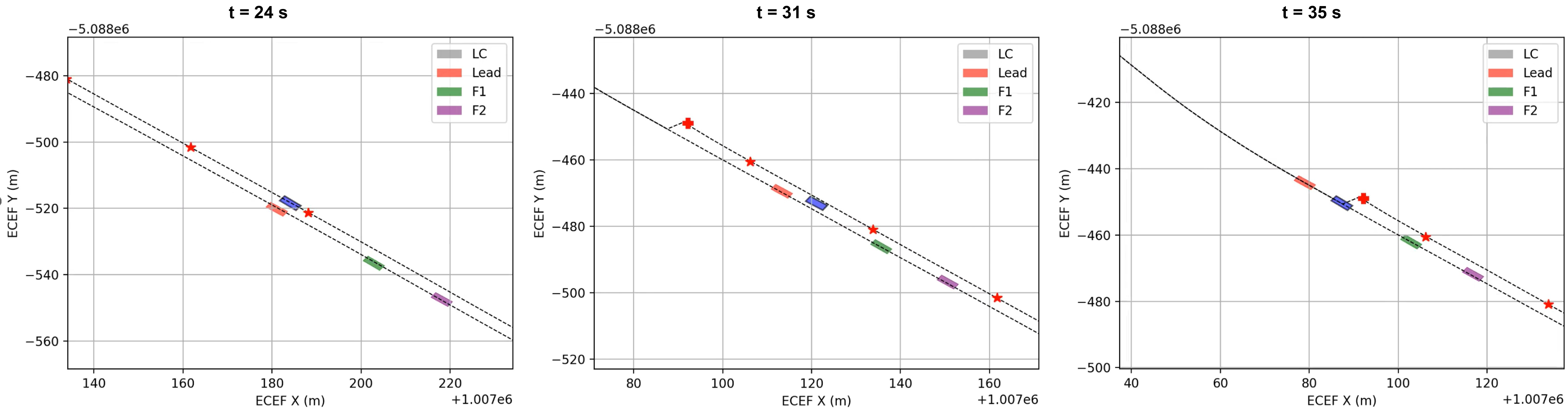}
    \caption{(Sample Example 1) Vehicle positions on the lane centerlines at three timestamps ($t = 24$, 31, and 35~s), shown left to right. The red ``asterisk'' indicates the right-turn pavement markings and the red ``plus'' sign indicates the end of the exclusive right-turn lane in the NB direction.}
    \label{fig:OverlayTrajectories}
\end{figure}}

\newcommand{\IndependentVariables}{
\begin{figure}[t]
    \centering
    \includegraphics[width=\linewidth]{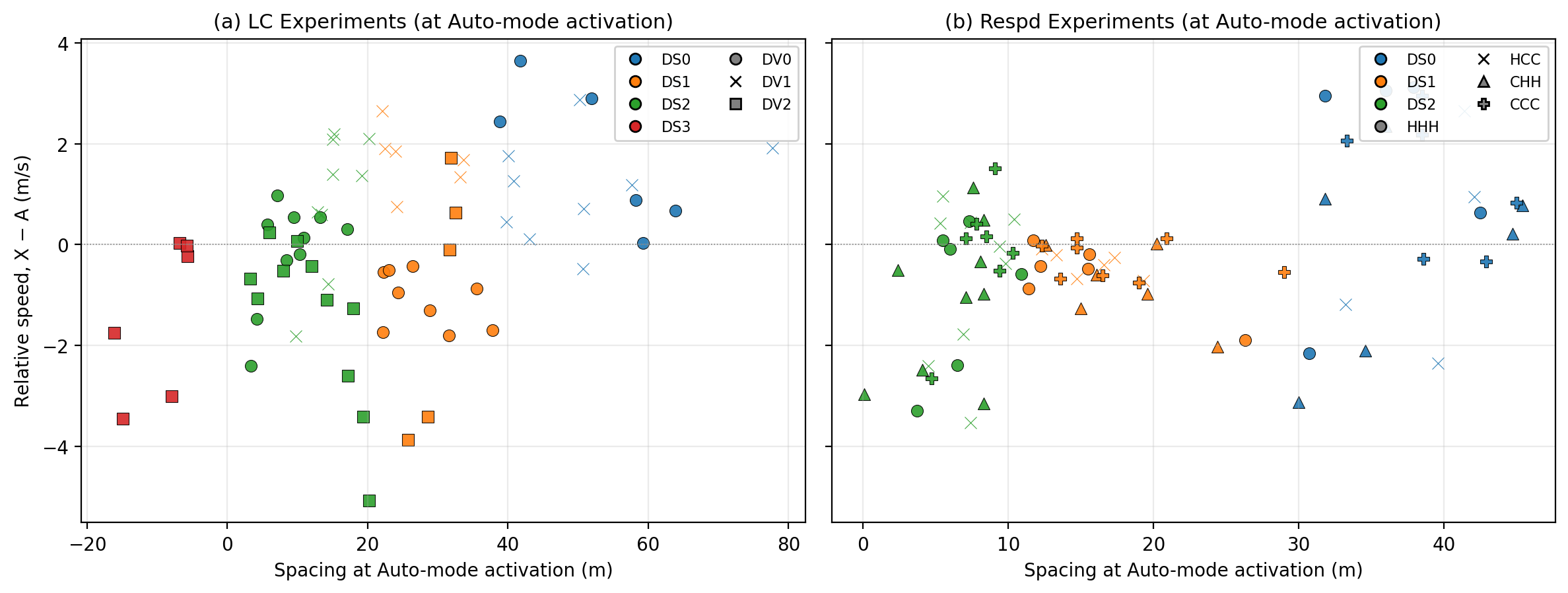}
    \caption{Sample distribution over the independent variables, measured at Auto-mode activation and coloured by the (relabeled) Discrete Spacing (DS) category. (a) LC experiments, with marker shape denoting the Discrete Velocity level (DV0--DV2); (b) Respd experiments (DV0 only), with marker shape denoting the driving-style configuration (HHH/HCC/CHH/CCC). The horizontal axis is the longitudinal spacing at activation and the vertical axis the relative speed of X with respect to the lead A.}
    \label{fig:IndependentVariables}
\end{figure}}

\newcommand{\SampleExampleLC}{
\begin{figure}[t]
    \centering
    \includegraphics[ height = 22cm]{Figures/SampleExampleLC.png}
    \caption{Sample Example 1 from LC Experiment (DS = 1, DV = 0)}
    \label{fig:Sample Example LC Experiments}
\end{figure}}

\newcommand{\OverlayTrajectoriesT}{
\begin{figure}[t]
    \centering
    \includegraphics[ height = 0.8 \paperheight]{Figures/OverlayedVehicleTrajectoryonLaneCenterlinesLC2.jpg}
    \caption{(Sample Example 2) Vehicles position and lane centerline at three timestamp t= 26 s, t = 33 s and t = 36 s. The red"asterisk" indicate the right-turn pavement markings and red "plus" sign indicates the end of exclusive right-turn lane in NB direction.}
    \label{fig:OverlayTrajectories 2}
\end{figure}}

\newcommand{\SampleExampleLCT}{
\begin{figure}[t]
    \centering
    \includegraphics[ height = 22cm]{Figures/SampleExampleLCExperiment2.png}
    \caption{Sample Example 2 from LC Experiment (DS = 2, DV = 0)}
    \label{fig:Sample Example LC Experiments 2}
\end{figure}}

\newcommand{\SampleExampleRespd}{
\begin{figure}[t]
    \centering
    \includegraphics[ height = 22cm]{Figures/SampleExample_Respd.png}
    \caption{Sample Example from Respd Experiment (DS = 0, Config: HHH)}
    \label{fig:Sample Example Respd Experiments}
\end{figure}}

\newcommand{\SampleExampleCombined}{
\begin{figure}[t]
    \centering
    \begin{subfigure}[t]{0.48\linewidth}
        \centering
        \includegraphics[width=\linewidth]{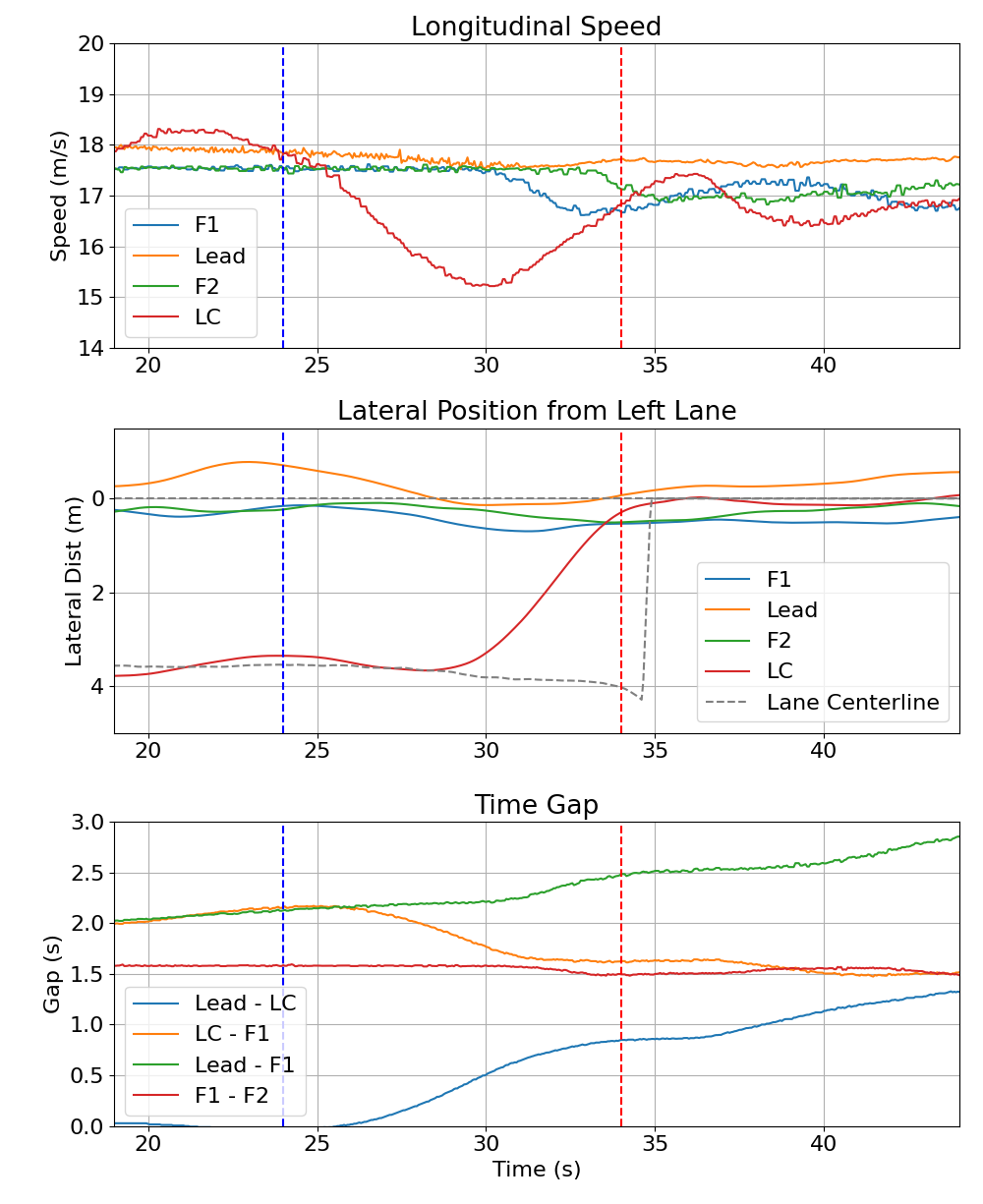}
        \caption{LC experiment ($DS=1$, $DV=0$)}
        \label{fig:Sample Example LC Experiments}
    \end{subfigure}\hspace{0.02\linewidth}
    \begin{subfigure}[t]{0.48\linewidth}
        \centering
        \includegraphics[width=\linewidth]{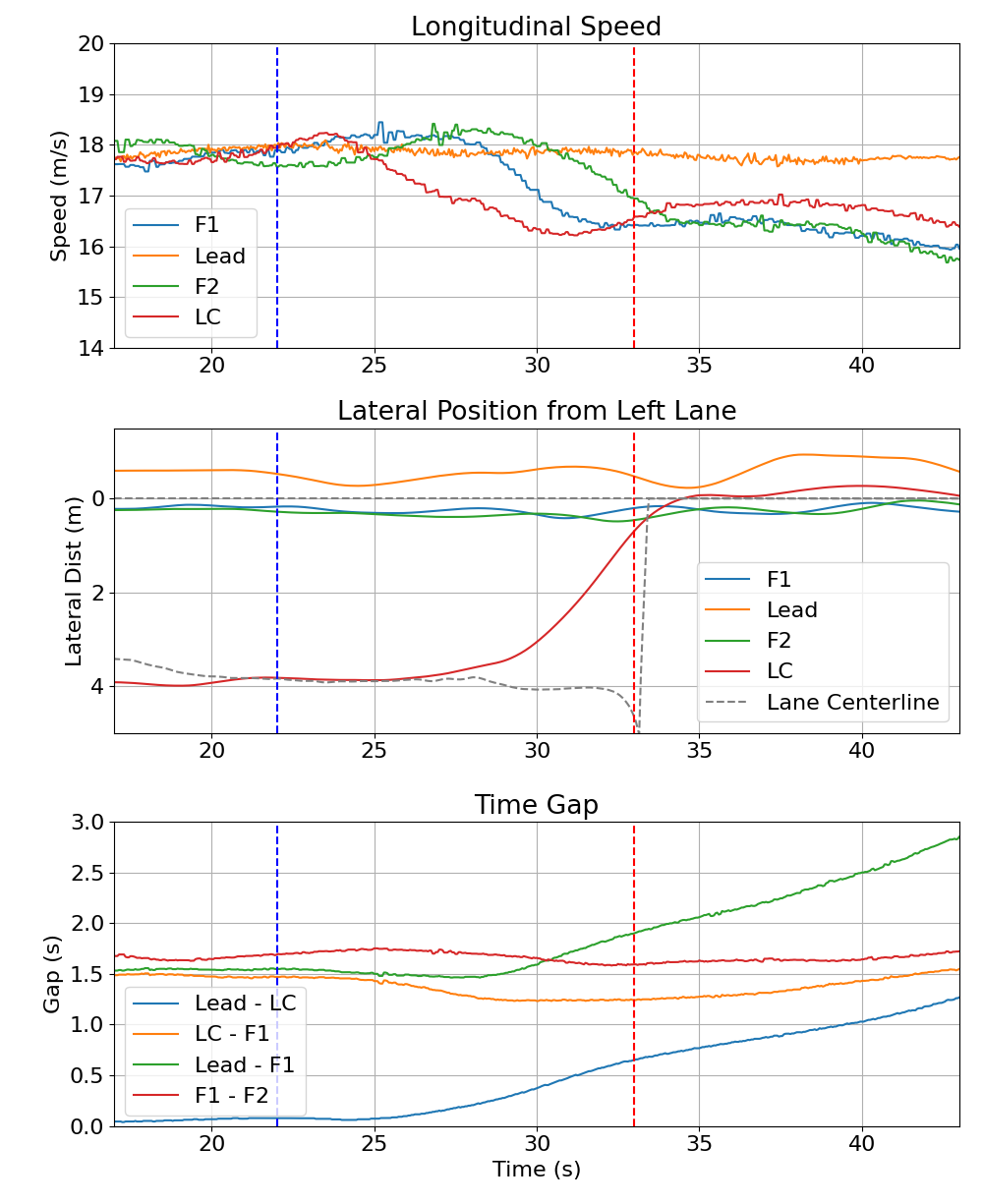}
        \caption{Respd experiment ($DS=0$, Config: HHH)}
        \label{fig:Sample Example Respd Experiments}
    \end{subfigure}
    \caption{Sample trajectories from (a) an LC-experiment trial and (b) a Respd-experiment trial. Each column shows, top to bottom, the longitudinal speed, lateral position (from the left-lane centerline), and time gap of the four vehicles. In every panel the blue vertical line marks Auto-mode activation ($t^{*}$) and the red vertical line marks merge completion.}
    \label{fig:Sample Example Combined}
\end{figure}}

\newcommand{\TargetVehiclesSpeedLC}{
\begin{figure}[t]
    \centering
    \includegraphics[width=\linewidth]{Figures/TargetVehiclesSpeedinLCExperiments.jpg}
    \caption{Target Vehicles Speed Distribution in LC Experiments (\textbf{Lead Vehicle}: Mean: 17.89 m/s Std Dev: 0.09 m/s Median: 17.87 m/s 25th \%: 17.81 m/s 75th \%: 17.97 m/s Min: 17.69 m/s Max: 18.10 m/s 
\textbf{F1 Vehicle}: Mean: 17.52 m/s Std Dev: 0.04 m/s Median: 17.53 m/s 25th \%: 17.50 m/s 75th \%: 17.55 m/s Min: 17.39 m/s Max: 17.62 m/s 
\textbf{F2 Vehicle}: Mean: 17.52 m/s Std Dev: 0.04 m/s Median: 17.53 m/s 25th \%: 17.51 m/s 75th \%: 17.55 m/s Min: 17.41 m/s Max: 17.58 m/s
)}
    \label{fig:Target Vehicles Speed in LC Experiments}
\end{figure}}

\newcommand{\TargetGapLC}{
\begin{figure}[t]
  \centering
  \includegraphics[width=\linewidth]{Figures/TargetGapinLCExperiment.jpg}
  \caption{Target Gaps in LC Experiments measured at Auto Mode Activation Time (\textbf{Lead–F1 Spacing:} Mean: 36.90 m Std Dev: 3.21 m Median: 36.10 m 25th \%: 35.30 m 75th \%: 37.33 m \textbf{}\textbf{F1–F2 Spacing:} Mean: 29.22 m Std Dev: 1.92 m Median: 28.75 m 25th \%: 27.98 m 75th \%: 29.88 m
)}
  \label{fig:Target Gap in LC Experiment}
\end{figure}}

\newcommand{\TargetGapRespd}{
\begin{figure}[t]
  \centering
  \includegraphics[width=\linewidth]{Figures/TargetGapinRespdExperiment.jpg}
  \caption{Target Gaps in Respd Experiments measured at Auto Mode Activation Time (\textbf{Lead–F1 Spacing}: Mean: 28.92 m Std Dev: 5.10 m Median: 27.65 m 25th \%: 25.35 m 75th \%: 30.75 m \textbf{F1–F2 Spacing}: Mean: 31.92 m Std Dev: 5.24 m Median: 31.30 m 25th \%: 28.55 m 75th \%: 34.15 m)
}
  \label{fig:Target Gap in Respd Experiment}
\end{figure}}

\newcommand{\TargetVehiclesSpeedRespd}{
\begin{figure}[t]
    \centering
    \includegraphics[width=\linewidth]{Figures/TargetVehiclesSpeedinRespdExperiments.jpg}
    \caption{Target Vehicles Speed Distribution in Respd Experiments (\textbf{Lead Vehicle}: Mean: 17.90 m/s Std Dev: 0.09 m/s Median: 17.91 m/s 25th \%: 17.82 m/s 75th \%: 17.98 m/s Min: 17.74 m/s Max: 18.12 m/s \textbf{F1 Vehicle}: Mean: 18.24 m/s Std Dev: 0.43 m/s Median: 18.34 m/s 25th \%: 17.97 m/s 75th \%: 18.49 m/s Min: 16.95 m/s Max: 18.93 m/s F2 Vehicle: Mean: 18.24 m/s Std Dev: 0.67 m/s Median: 18.21 m/s 25th \%: 17.78 m/s 75th \%: 18.49 m/s Min: 16.84 m/s Max: 19.84 m/s
}
    \label{fig:Target Vehicles Speed in Respd Experiments}
\end{figure}}

\newcommand{\GapAcceptance}{
\begin{figure}[t]
    \centering
    \includegraphics[width=\linewidth]{Figures/GapAcceptance.jpg}
    \caption{Accepted Gap across independent variables}
    \label{fig:Gap Acceptance}
\end{figure}}

\newcommand{\LeadLagEvolution}{
\begin{figure}[!htbp]
    \raggedright
    \captionsetup{justification=raggedright,singlelinecheck=false}
    \includegraphics[width=\linewidth]{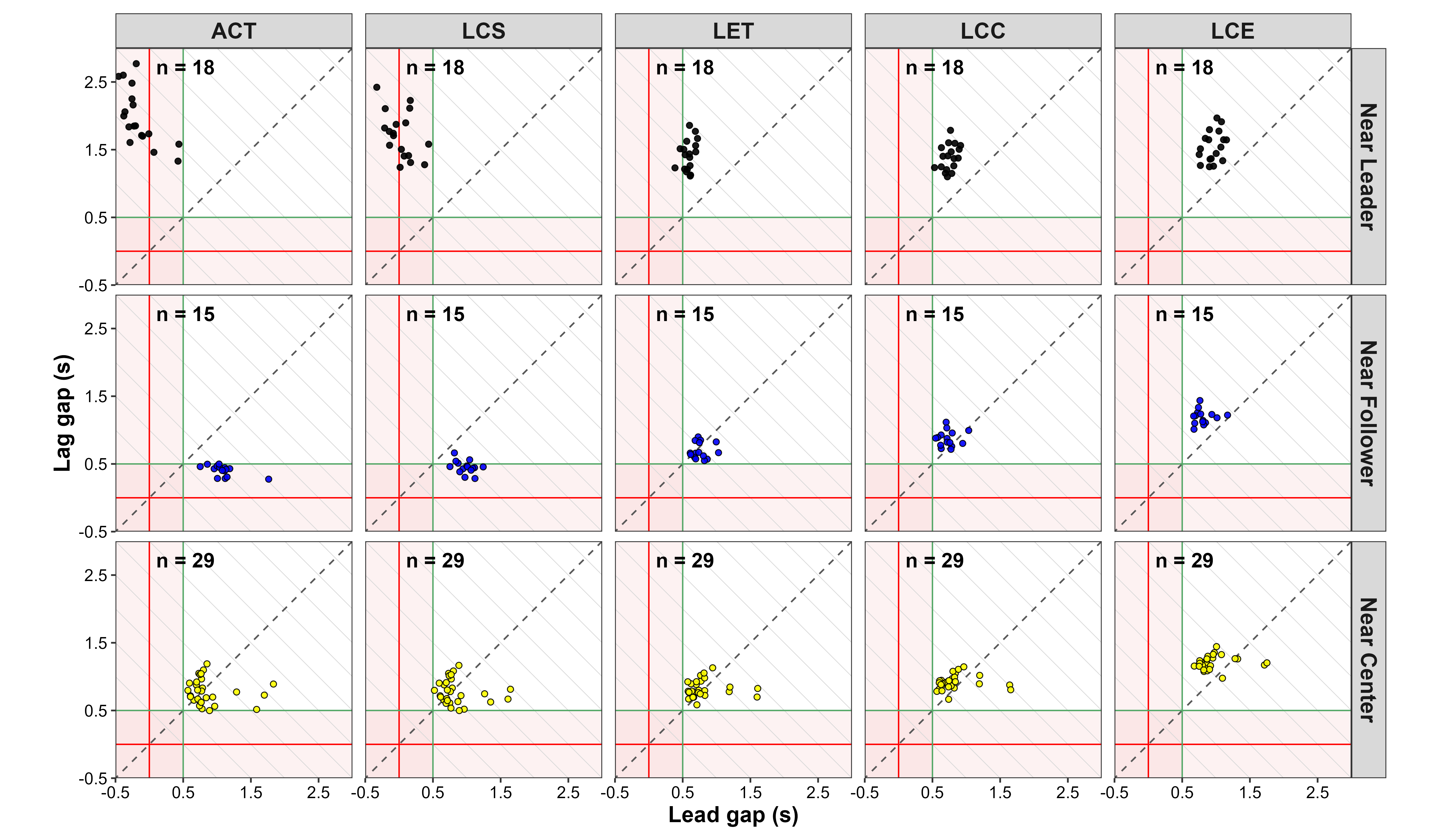}
    \caption{Lead--lag evolution throughout LC in tAV.}
    \label{fig:lead-lag-evolution}
\end{figure}}

\newcommand{\LeadLagDistribution}{
\begin{figure}[!htbp]
    \raggedright
    \captionsetup{justification=raggedright,singlelinecheck=false}
    \includegraphics[width=\linewidth]{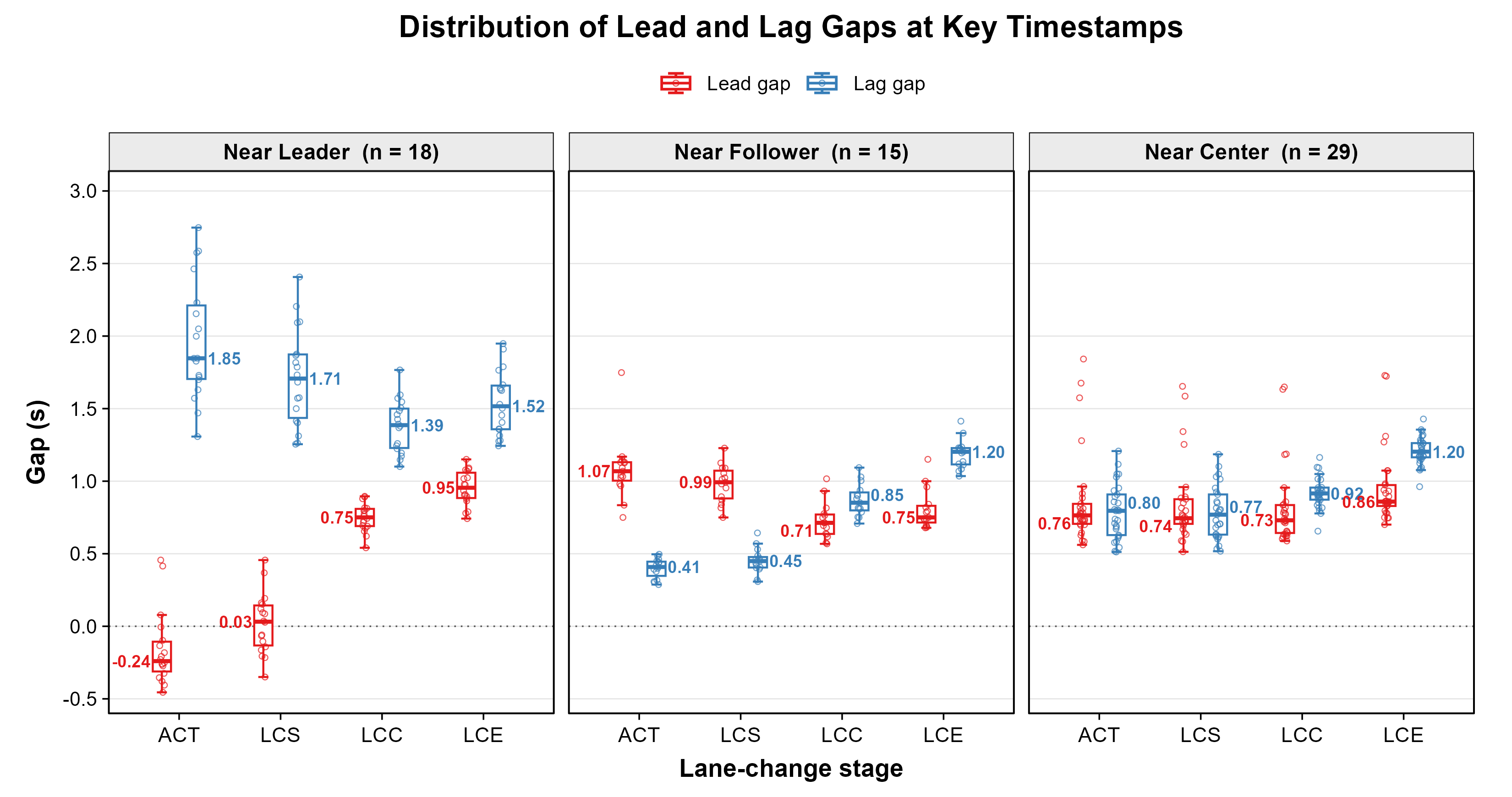}
    \caption{Distribution of lead and lag gaps in key time stamps in tAV.}
    \label{fig:gap-distributions}
\end{figure}}

\newcommand{\TotalGapDistribution}{
\begin{figure}[!htbp]
    \raggedright
    \captionsetup{justification=raggedright,singlelinecheck=false}
    \includegraphics[width=\linewidth]{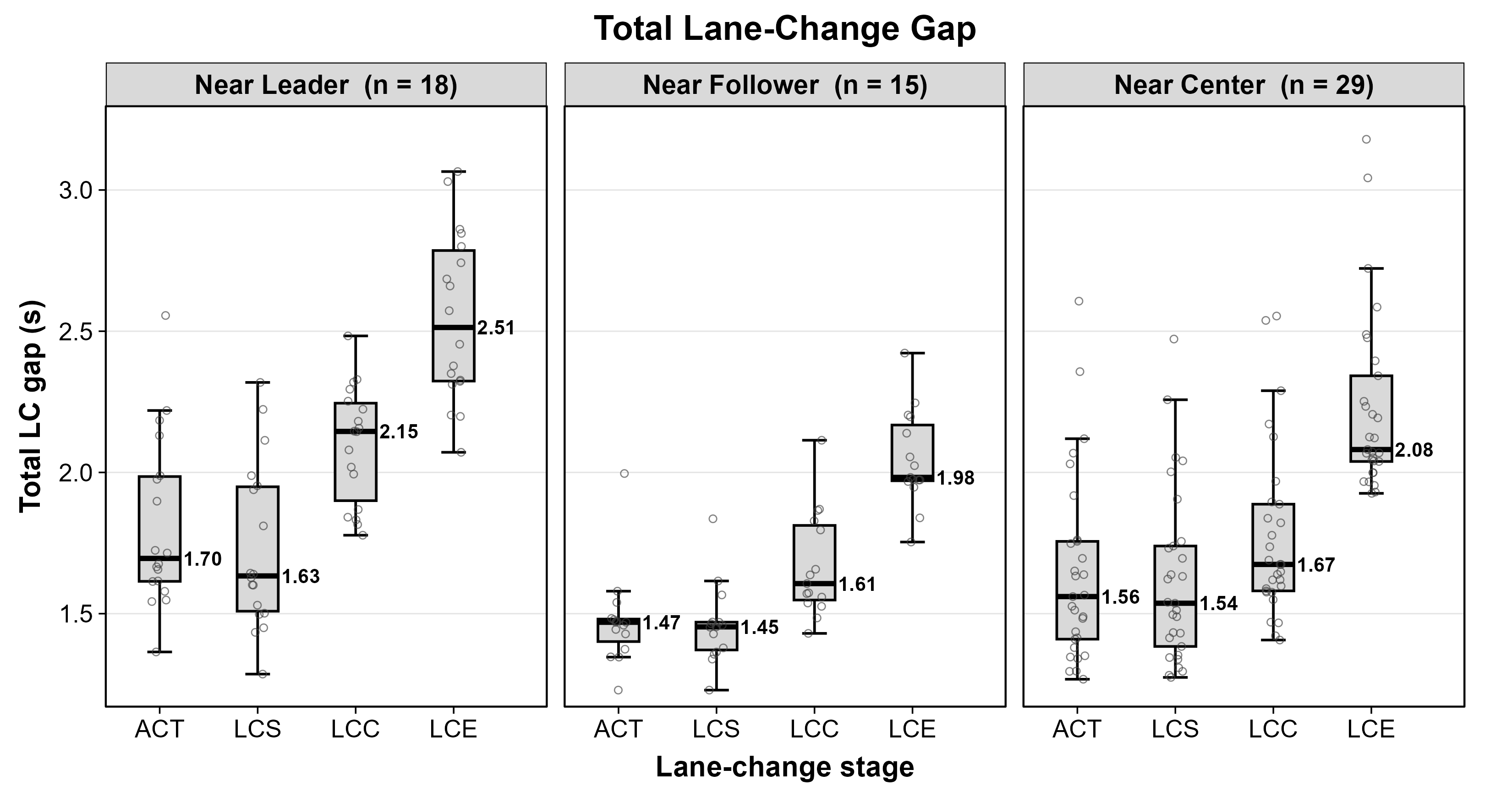}
    \caption{Distribution of LC gaps in key time stamps in tAV.}
    \label{fig:total-gap-distribution}
\end{figure}}

\newcommand{\RiskYOneTwo}{
\begin{figure}[!htbp]
    \centering

    \begin{minipage}[t]{0.48\linewidth}
        \centering
        \includegraphics[
            width=\linewidth,
            height=5cm,
            keepaspectratio
        ]{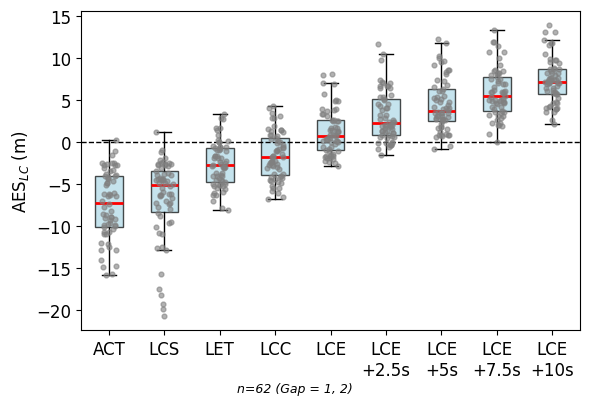}
        \caption{AES$_{LC}$ plot from ACT to LCE+10s; Gap = 1, 2 ($n = 62$).}
        \label{fig:Y1}
    \end{minipage}
    \hfill
    \begin{minipage}[t]{0.48\linewidth}
        \centering
        \includegraphics[
            width=\linewidth,
            height=5cm,
            keepaspectratio
        ]{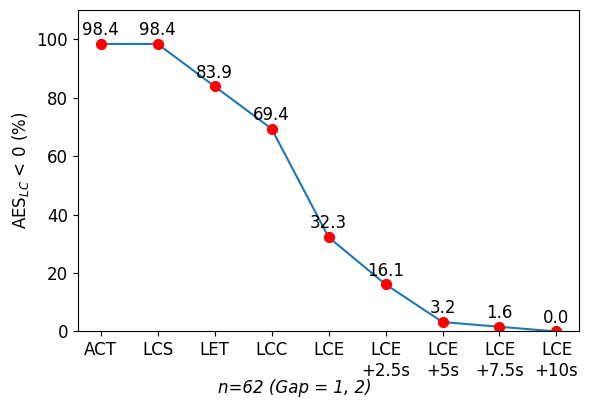}
        \caption{Fraction of at-risk cases ($AES_{LC}<0$); Gap = 1, 2 ($n = 62$).}
        \label{fig:Y2}
    \end{minipage}

\end{figure}
}

\newcommand{\RiskYThree}{
\begin{figure}[!htbp]
    \raggedright
    \captionsetup{justification=raggedright,singlelinecheck=false}
    \includegraphics[
        width=\linewidth]{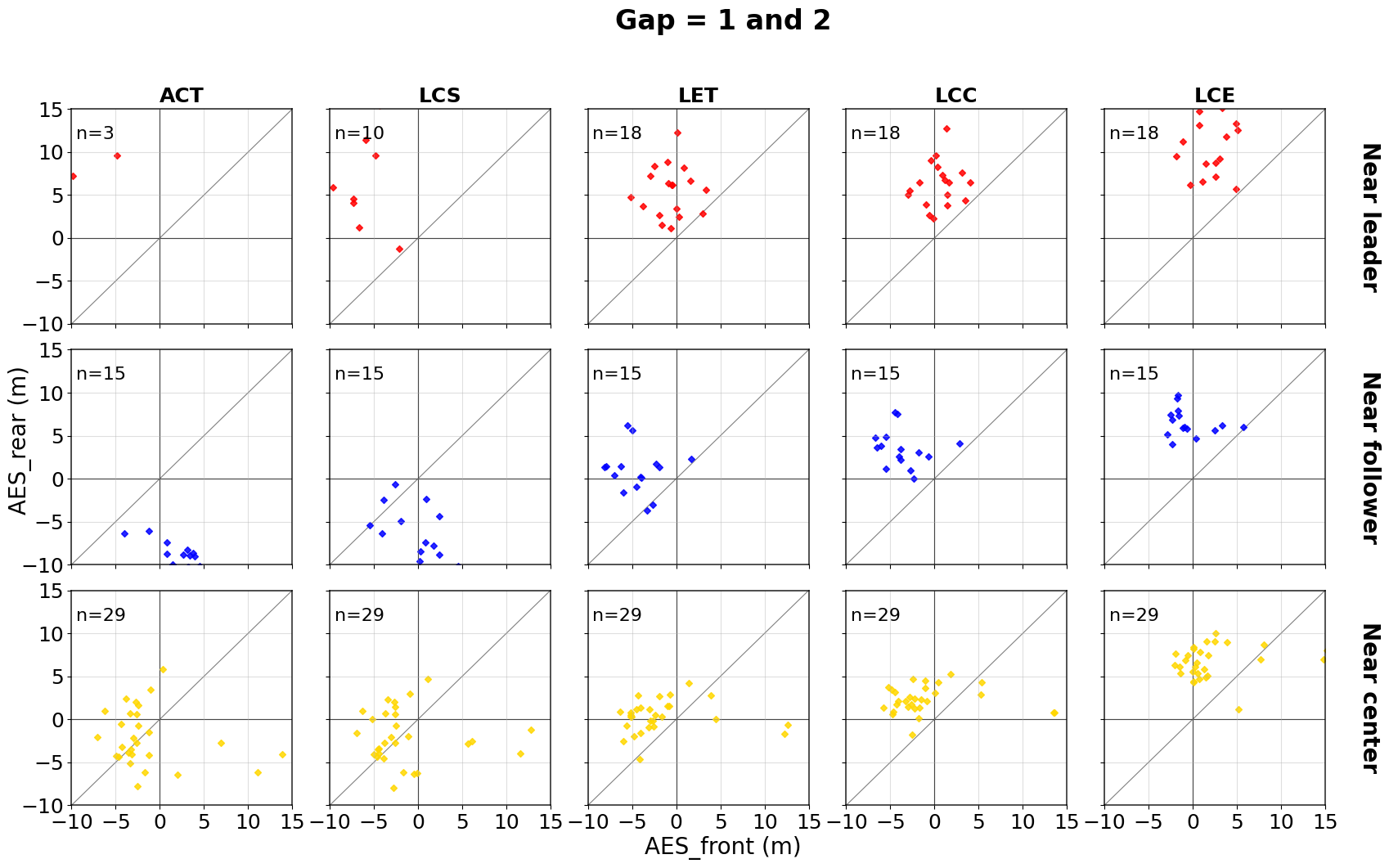}
    \caption{Risk source; $n = 62$, Gap = 1, 2.}
    \label{fig:Y3}
\end{figure}}

\newcommand{\RiskYFourFive}{
\begin{figure}[!htbp]
    \centering

    \begin{minipage}[t]{0.48\linewidth}
        \centering
        \includegraphics[
            width=\linewidth,
            height=5cm,
            keepaspectratio
        ]{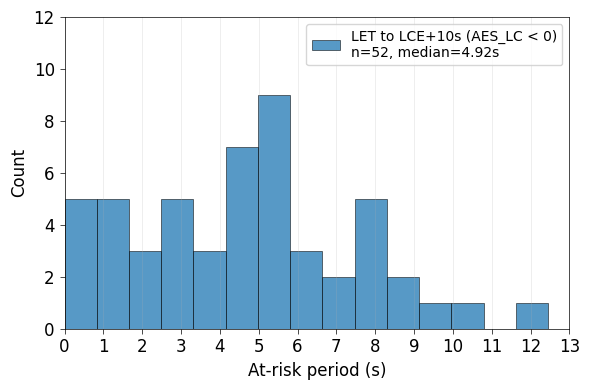}
        \captionof{figure}{Distribution of at-risk period between LET and LET+10s; Gap = 1, 2 ($n = 62$).}
        \label{fig:Y4}
    \end{minipage}
    \hfill
    \begin{minipage}[t]{0.48\linewidth}
        \centering
        \includegraphics[
            width=\linewidth,
            height=5cm,
            keepaspectratio
        ]{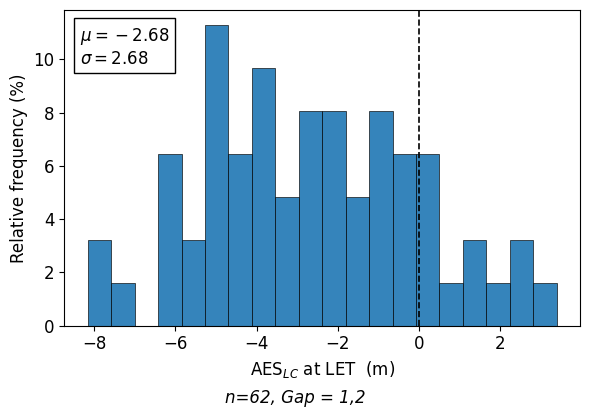}
        \captionof{figure}{AES$_{LC}$ distribution at LCE; Gap = 1, 2 ($n = 62$).}
        \label{fig:Y5}
    \end{minipage}

\end{figure}
}

\newcommand{\KeyTimestampFigure}{
\begin{figure}[!htbp]
    \raggedright
    \includegraphics[width=0.8\textwidth]
    {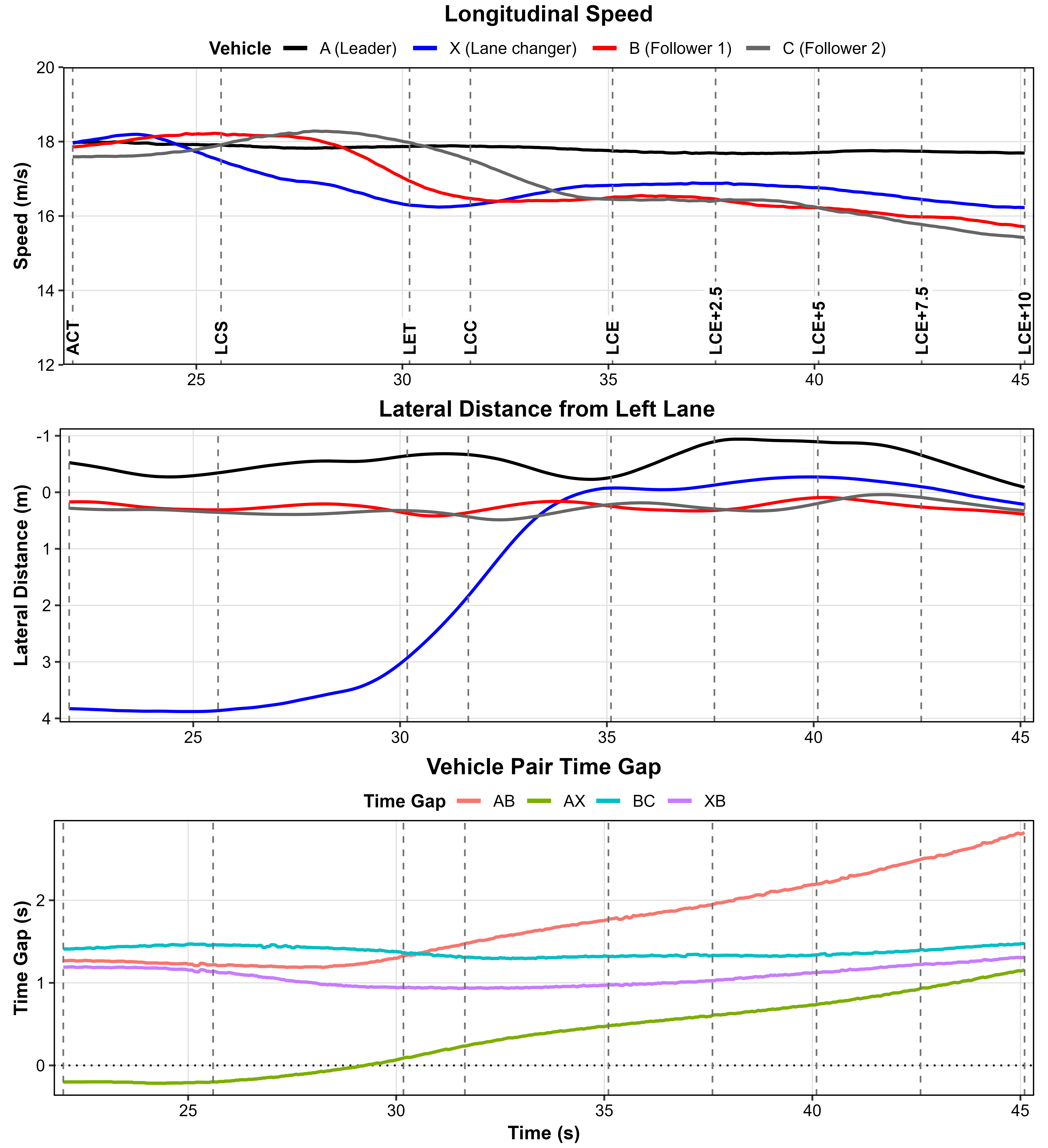}
    \caption{Example trajectories of the four experimental vehicles during a mandatory lane change. The panels show the longitudinal speed, lateral distance from the left lane, and vehicle-pair time gaps, respectively. The vertical dashed lines indicate the key lane-change timestamps: automation activation (ACT), lane-change start (LCS), left-edge touching (LET), lane-change crossing (LCC), lane-change end (LCE), and the post-lane-change timestamps at 2.5, 5, 7.5, and 10~s after LCE.}
    \label{fig:key-timestamp-example}
\end{figure}
}
\newcommand{\VehicleInstrumentation}{
\begin{table}[t]
\footnotesize
\centering
\caption{Vehicle Specifications and Instrumentation Summary}
\label{tab:vehicle_instrumentation}
\renewcommand{\arraystretch}{1.2}
\resizebox{\textwidth}{!}{
\begin{tabular}{>{\raggedright\arraybackslash}p{2.3cm} >{\raggedright\arraybackslash}p{3.4cm} >{\raggedright\arraybackslash}p{2.7cm} >{\raggedright\arraybackslash}p{3.4cm} >{\raggedright\arraybackslash}p{3.2cm} >{\centering\arraybackslash}p{1.4cm}}
\toprule
\rowcolor{gray!20}
\textbf{Vehicle Type} & \textbf{System} & \textbf{Sensor / Device} & \textbf{Model} & \textbf{Accuracy (RMS)} & \textbf{Freq.} \\ 
\midrule

\multirow{2}{*}{A (Lead)}
  & \multirow{2}{=}{Cruise control}
  & INS & NovAtel Dual-Antenna RTK-GPS & 
  \begin{tabular}[t]{@{}l@{}}
    Position: 0.01 m\\
    Heading: 0.09° @ 2 m \\
    Velocity: 0.02 m/s
  \end{tabular} & 20 Hz \\
  & & & & & \\

\midrule

\multirow{3}{*}{X \& B}
  & \multirow{3}{=}{ tAV}
  & INS  & Inertial Labs INS-DM (RTK) & 
  \begin{tabular}[t]{@{}l@{}}
    Position: 0.01 m\\
    Velocity: 0.01 m/s\\
    Heading: 0.05°
  \end{tabular} & 20 Hz \\
  & & Cameras & 4× Sony ZV-1F & 1920×1080, 84° FOV & 30 fps \\
  & & CAN Device & VLinker FS Bluetooth & N/A & N/A \\

\midrule

\multirow{3}{*}{C}
  & \multirow{3}{=}{tAV}
  & INS & Inertial Labs INS-DU (RTK) & 
  \begin{tabular}[t]{@{}l@{}}
    Position: 0.01 m\\
    Velocity: 0.05 m/s\\
    Heading: 0.10°
  \end{tabular} & 20 Hz \\
  & & Cameras & 2× Sony ZV-1F & 1920×1080, 84° FOV & 30 fps \\
  & & CAN Device & VLinker FS Bluetooth & N/A & N/A \\

\bottomrule
\end{tabular}}
\end{table}
}
\newcommand{\ParameterSettings}{
\begin{table}[t]
\centering
\caption{Driving Settings (LC and Respd Experiments)}
\label{tab:lc_respd_settings}
\renewcommand{\arraystretch}{1.2}
\begin{tabular}{|p{4.1cm}|p{2.4cm}|p{2.4cm}|p{2.4cm}|p{2.4cm}|}
\hline
\rowcolor{red!20}
\textbf{Parameters} & \textbf{X} & \textbf{A} & \textbf{B} & \textbf{C} \\
\hline
\multicolumn{5}{|c|}{\cellcolor{gray!15}\textbf{LC Experiment}} \\
\hline
Driving Mode & Auto & Cruise control\textsuperscript{1} & ACC & ACC \\
Driving Style & Hurry & NA & Headway 2\textsuperscript{2} & Headway 2\textsuperscript{2} \\
Set Speed & Speed limit & 40 mph & 40 mph & 40 mph \\
Speed Offset & 40\%\textsuperscript{3} & NA & 0\% & 0\% \\
\hline
\multicolumn{5}{|c|}{\cellcolor{gray!15}\textbf{Respd Experiment}} \\
\hline
Driving Mode & Auto & Cruise control & Auto & Auto \\
Driving Style & Variable\textsuperscript{4} & NA & Variable & Variable \\
Set Speed & Speed limit & 40 mph & Speed limit & Speed limit \\
Speed Offset & Variable & NA &  & Variable \\
\hline
\end{tabular}
\vspace{1em}

\begin{flushleft}
\small
\textsuperscript{1} Vehicle A (lead) runs cruise control (constant set speed) and is not a tAV. \\
\textsuperscript{2} Headway 2: Minimum Headway in ACC mode \\
\textsuperscript{3} Speed Offset (from detected speed limit): 20\% for Chill Mode and 40\% for Hurry Mode \\
\textsuperscript{4} Variable: Either \textit{Hurry} or \textit{Chill} mode per trial. (see configurations HHH/HCC/CHH/CCC) \\
\end{flushleft}
\end{table}
}

\newcommand{\SampleSize}{
\begin{table}[t]
\centering
\footnotesize
\renewcommand{\arraystretch}{1.1}
\caption{Sample size distribution: (a) LC experiments by DS$\times$DV ($n=72$) and (b) Respd experiments by driving-style configuration $\times$ DS ($n=78$). Empty cells denote combinations that were not sampled.}
\label{tab:sample_distribution_combined}
\begin{minipage}[t]{0.45\linewidth}
\centering
\textbf{(a) LC: DS $\times$ DV}\\[3pt]
\begin{tabular}{|c|ccc|c|}
\hline
\rowcolor{red!10}
\textbf{DS} & \textbf{DV0} & \textbf{DV1} & \textbf{DV2} & \textbf{Tot.} \\
\hline
0 & 6  & 10 &    & 16 \\
1 & 9  & 6  & 5  & 20 \\
2 & 10 & 9  & 11 & 30 \\
3 &    &    & 6  & 6  \\
\hline
\textbf{Tot.} & 25 & 25 & 22 & \textbf{72} \\
\hline
\end{tabular}
\end{minipage}\hfill
\begin{minipage}[t]{0.52\linewidth}
\centering
\textbf{(b) Respd: Config $\times$ DS}\\[3pt]
\begin{tabular}{|c|ccc|c|}
\hline
\rowcolor{red!10}
\textbf{Config} & \textbf{DS0} & \textbf{DS1} & \textbf{DS2} & \textbf{Tot.} \\
\hline
HHH & 5 & 6 & 6  & 17 \\
HCC & 4 & 6 & 9  & 19 \\
CHH & 6 & 6 & 9  & 21 \\
CCC & 6 & 8 & 7  & 21 \\
\hline
\textbf{Tot.} & 21 & 26 & 31 & \textbf{78} \\
\hline
\end{tabular}
\end{minipage}
\end{table}
}

\section{Introduction}

Vehicles operating beyond conventional SAE Level 1--2 automation but short of full autonomy are increasingly present on public roadways \cite{Ammourah2024IntroductionExtraction}. This study focuses on these vehicles, referred to as transitional autonomous vehicles (tAVs). What distinguishes tAVs from the Level 1--2 systems that preceded them is not an incremental improvement in longitudinal control, but the addition of a qualitatively new capability: automated lane changing. Unlike driver-initiated assisted lane changing, in which the driver makes the lane-change decision and the automation primarily executes the maneuver, a tAV can independently conduct the complete lane-change, including both lane-change decision making and maneuver execution, while remaining under driver supervision. We refer to these systems as transitional autonomous vehicles (tAVs) to distinguish them from conventional SAE Level 2 driver-assistance functions. Lane changing thus moves from a task the driver performs to a task the vehicle does that is already among the most safety-critical in everyday driving---accounting for roughly 4.2\% of police-reported crashes in the United States \cite{NHTSA2022SummarySystems}-This behavior of tAVs has not yet been sufficiently characterized. Beyond crash risk, lane changing requires tactical decisions such as gap selection under time pressure, and its effects extend to traffic capacity as well as safety, contributing to disturbances and bottlenecks even in the absence of a collision \cite{Cassidy1999SomeBottlenecks,Shi2021EmpiricalSettings,Zheng2014RecentChanging}.

Lane-change capabilities are already deployed on public roads, often under functions described as ``Automatic Lane Change,'' including Tesla's Full Self-Driving (Supervised) \cite{TeslaInc.2025ModelSupervised}, GM Super Cruise with hands-free driving \cite{GeneralMotors2025HowCruise}, and Ford BlueCruise 1.5 \cite{FordMotorCompany2025BlueCruise}. These systems can initiate lane changes based on the driving environment without driver confirmation. Automated lane changing is therefore already present in daily traffic streams. The resulting lateral behavior may differ from that of both human-driven vehicles (HDVs) and Level 4 self-driving vehicles (SDCs), which operate without continuous driver supervision within their operational design domains.

Despite their growing deployment, empirical studies of automated vehicle (SAE Level 2 and above) lane-changing behavior remain scarce. Existing evidence primarily comes from three sources. First, the availability of automated vehicle trajectories through the Waymo Open Motion Dataset (WOMD) has enabled researchers to investigate behavioral differences between automated and human-driven vehicles \cite{Waymo2025WaymoDataset}. The dataset captures Level 4 automated vehicle operations under real-world traffic conditions. Analyses of these data have shown that Level 4 automated vehicles adopt larger lead and lag gaps, longer lane-change execution times, and lower acceleration variability than human drivers \cite{Ali2024InvestigatingDataset}, that following vehicles experience lower speed volatility and reduced crash risk \cite{Wen2023AnalysisData}, and that automated vehicle lane changes can still provoke abrupt responses from surrounding human drivers, increasing rear-end conflict potential \cite{Ali2025AutonomousFollower}.

Second, a Third Generation Simulation (TGSIM) dataset captures Level 1--3 automated vehicles operating in naturalistic highway and urban environments using aerial and infrastructure-based videography \cite{Ammourah2024IntroductionExtraction}. This dataset includes 32 SAE Level 2 assisted lane-change events collected from the I-90/I-94 and I-294 corridors \cite{Ammourah2024IntroductionExtraction, FHWA2024ThirdFHWA-JPO-24-133}. Although it contains mandatory lane-change scenarios, published studies have not yet provided an empirical characterization of transitional autonomous vehicle (tAV) lane-changing behavior. Existing analyses have instead focused primarily on discretionary lane changes while pooling Level 1 and Level 2 vehicles, rather than isolating mandatory lane-changing behavior of tAVs \cite{Zhang2025InvestigationDataset}.

Third, the Joint Research Centre (JRC) experimental campaign conducted by Mattas et al.~\cite{Mattas2025SafetyExperiments} comes closest to the present study. Six commercially available SAE Level 2 vehicles equipped with assisted lane-change functionality were evaluated on Interstate 24 near Nashville, Tennessee, under a range of operating modes, including manual, driver-requested, driver-confirmed, system-initiated, and interrupted lane changes. The study found that each system exhibited consistent lane-change behavior, with four of the five vehicles performing lane changes at lower speeds and decelerations than human drivers, while one completed the maneuver more assertively in approximately five seconds. For safety assessment, the authors analyzed 161 cases in which the automated vehicle merged ahead of a target-lane follower within 100~m. Manual and assisted lane changes frequently began with follower--lane-changer distances below the minimum values specified by UN Regulations 79 and 171, whereas assisted lane changes generally induced only moderate deceleration in the following vehicle, with observed deceleration magnitudes being smaller than simulation predictions. Notably, only ten reported lane changes were system initiated, and these cases were excluded from the safety-margin analysis. Furthermore, the dataset has not yet been released publicly.

Collectively, these studies provide valuable evidence on Level 4 AV lane changing and driver-initiated assisted lane-change systems. However, two research gaps remain. First, systematic empirical data remain lacking for tAVs that independently conduct the complete lane-change process, including both lane-change decision making and maneuver execution. Second, no empirical study has yet characterized tAV lane-changing behavior or examined its potential impacts on traffic flow and safety.

To address these gaps, we developed the North Carolina Transitional Autonomous Vehicle Lane-Changing (NC-tALC) dataset \cite{Sharma2026IntroducingExperimentsb} through controlled mandatory lane-change experiments conducted on public roads. The experiments systematically vary the initial lane-change conditions, including relative spacing and relative speed, while using relatively small candidate target gaps to create challenging lane-change scenarios for the tAV. Additionally, two other tAVs in the target lane serve as followers, allowing tAV behavior to be observed both for the lane-changing vehicle and for responding followers. Using this dataset, we characterize tAV behavior throughout the complete lane-change process. The analysis focuses on two behavioral features: (i) the evolution and convergence of lead--lag gaps and (ii) the evolution and source of potential collision risk. To the best of our knowledge, the NC-tALC dataset provides an empirical foundation that is unavailable in existing AV lane-changing studies, while enabling new behavioral and safety insights from observed vehicle trajectories.

The remainder of the paper is organized as follows. Section~\ref{sec:experiments} describes the experimental design. Section~\ref{sec:data_method} presents the data processing procedures and analytical methods. Sections~\ref{sec:gap-evolution} and~\ref{section:Risk Assessment} present the findings on lead--lag gap evolution and collision risk assessment, respectively. Finally, Section~\ref{sec:conclusion} concludes the paper and discusses its implications.

\section{Experiments}
\label{sec:experiments}

This section introduces the experimental study, field setup, initial-condition control, and resulting sample used in the subsequent analyses.

\subsection{Experimental Study and Scope}
\label{sec:experimental_scope}

The NC-tALC dataset includes two controlled mandatory lane-change (LC) experiments that differ in the automation mode of the target-lane followers. The NC-tAV--tAV Follower experiment includes a lane changer and two potential target-lane followers operating as transitional automated vehicles (tAVs). The NC-tAV--ACC Follower experiment uses the same experimental setup, except that the two target-lane followers operate in adaptive cruise control (ACC) mode.

This paper focuses on analyzing the NC-tAV--tAV Follower experiment, while studying the NC-tAV--ACC Follower experiment is reserved for a separate paper. Throughout the remainder of this paper, the term \textit{tAV dataset} refers to the data collected from the NC-tAV--tAV Follower experiment. This section summarizes the experimental design and the data required for the subsequent analyses; additional technical details are provided in a separate experimental report.

\subsection{Experiment Scenario, Vehicle Configuration, and Field Control}
\label{sec:scenario_configuration}

\begin{figure*}[!t]
    \centering

    \begin{subfigure}[t]{0.48\textwidth}
        \centering
        \includegraphics[width=\linewidth]{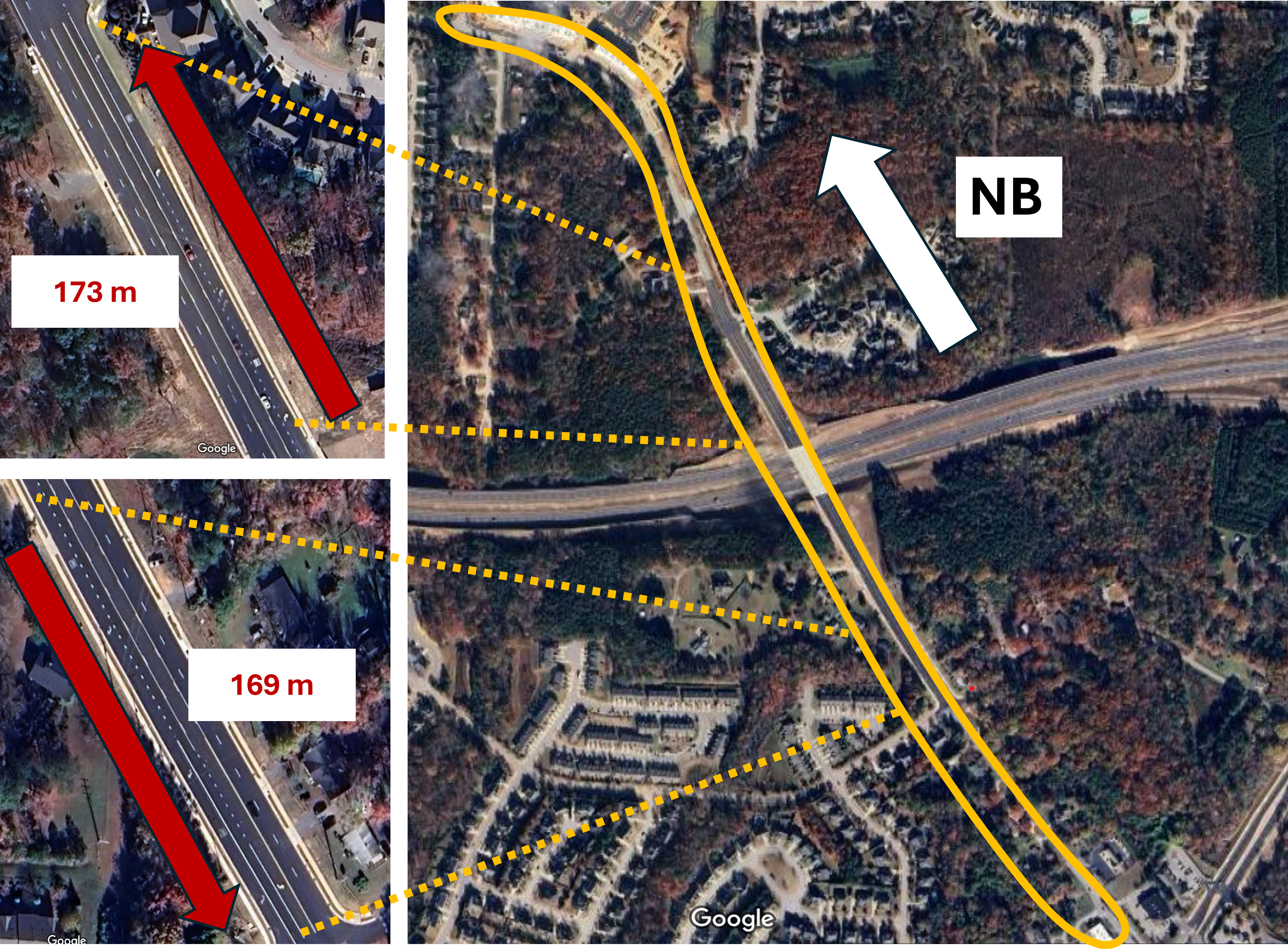}
        \caption{Test-site location and surrounding roadway network.}
        \label{fig:test_site_location}
    \end{subfigure}
    \hfill
    \begin{subfigure}[t]{0.48\textwidth}
        \centering
        \includegraphics[width=\linewidth]{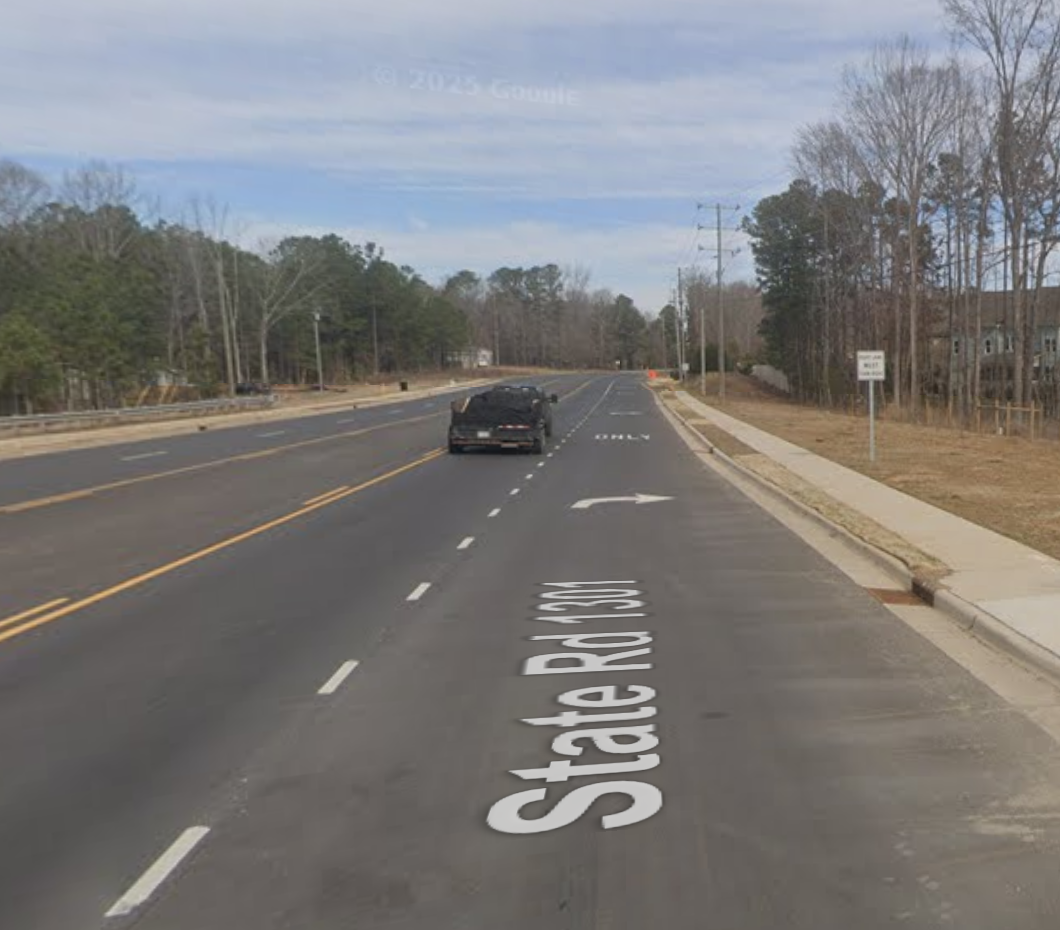}
        \caption{Field view of the experimental roadway segment.}
        \label{fig:test_site_field}
    \end{subfigure}

    \vspace{0.5em}

    \begin{subfigure}[t]{0.82\textwidth}
        \centering
        \includegraphics[width=\linewidth]{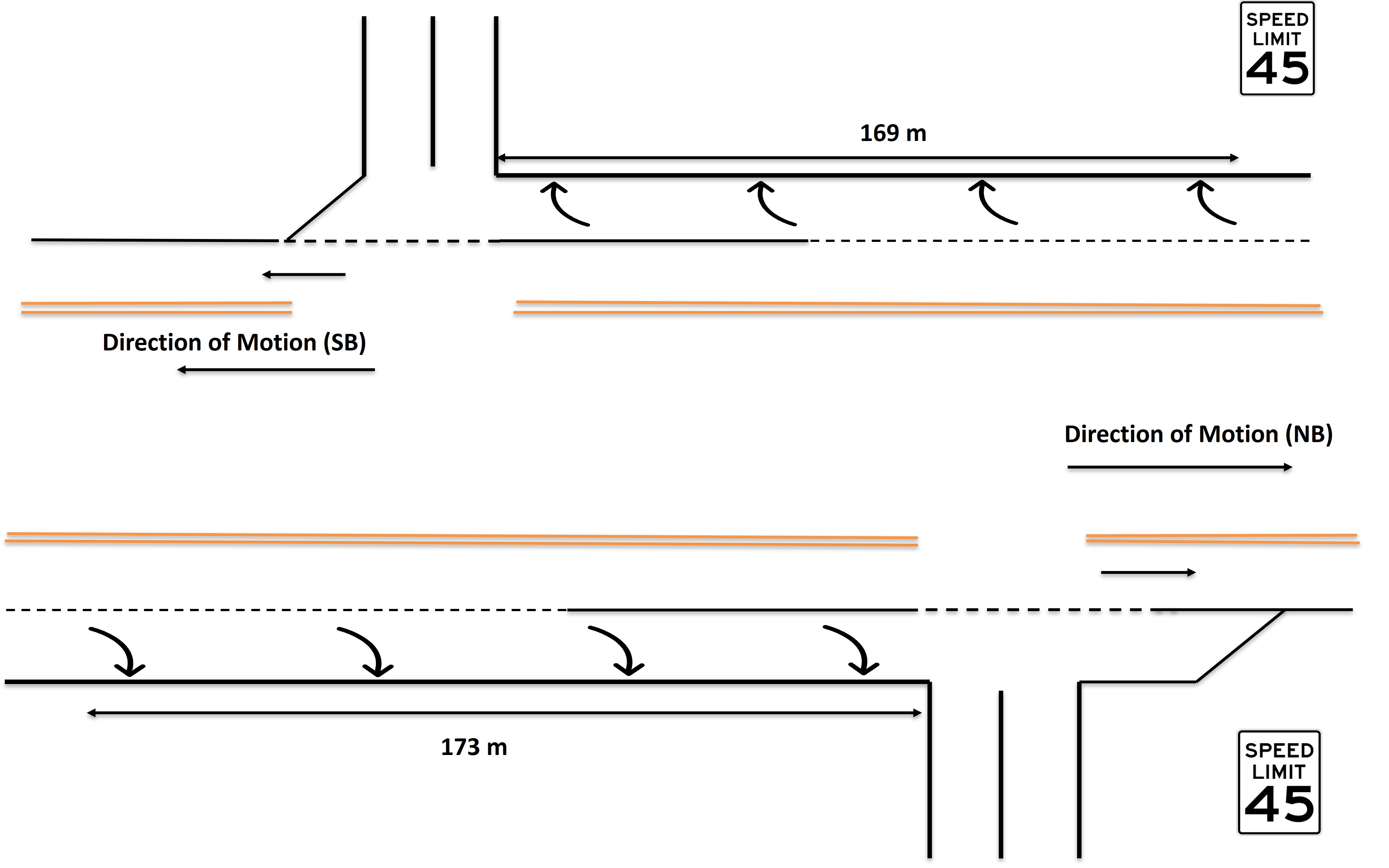}
        \caption{Roadway geometry and mandatory lane-change test layout.}
        \label{fig:test_site_layout}
    \end{subfigure}

    \caption{Experimental site on Sunset Lake Road in Apex, North Carolina: 
    (a) test-site location, (b) field environment, and (c) roadway and experimental layout.}
    \label{fig:test_site_combined}
\end{figure*}

The experiments were conducted during the daytime in the summer of 2025 on Sunset Lake Road in Apex, North Carolina. The selected site provides comparable northbound and southbound roadway geometries, with exclusive right-turn-lane lengths of approximately 173~m and 169~m, respectively. The roadway has a posted speed limit of 45~mph, predominantly straight segments, and a negligible grade (less than 2\%). Figure~\ref{fig:test_site_combined} presents the test-site location, field environment, and experimental roadway layout.

In both travel directions, a physical buffer is located downstream of the exclusive right-turn-lane endpoint, beyond which the pavement gradually tapers and its usable width decreases to zero. The lane-changing transitional automated vehicle (tAV) was assigned a destination that required it to exit the exclusive right-turn lane and merge into the adjacent left lane, thereby creating a mandatory lane-change scenario. Testing was conducted under clear and sunny conditions on a public roadway with live traffic and without lane closures. The roadway geometry, weather, visibility, and traffic conditions were within the operational design domain of the tested tAVs.

Four vehicles were used in each trial, denoted as X, A, B, and C, as illustrated in Figure~\ref{fig:Vehicle Configuration}. Vehicle A was a 2019 Lincoln MKZ equipped with adaptive cruise control (ACC), while vehicles X, B, and C were produced by the same manufacturer and had the same software version.
% TODO: Confirm the model and model year of Vehicle A.
% TODO: Confirm the manufacturer, vehicle models, and software version of X, B, and C.

Vehicles X, B, and C could operate either in ACC mode or in transitional automated vehicle (tAV) mode. In tAV mode, the automated driving system made driving decisions and executed longitudinal and lateral maneuvers under continuous human-driver supervision. Vehicle X served as the lane-changing tAV and traveled in the exclusive right-turn lane before performing the mandatory lane change into the adjacent left lane. Vehicles A, B, and C formed a three-vehicle platoon in the target lane. Vehicle A operated under cruise control, while B and C operated under tAV control with human-driver supervision.

The four vehicles maintained their assigned experimental roles throughout the trials: A as the target-lane leader, B and C as the first and second target-lane followers, respectively, and X as the lane-changing vehicle.
% TODO: Confirm that the same physical vehicles maintained the A, B, C, and X roles in every trial.

This four-vehicle arrangement provided multiple candidate target gaps for X and enabled its target-gap selection and subsequent lane-change execution to be examined under controlled initial conditions.

% Existing vehicle-configuration figure from the previous draft.
% It will be numbered automatically as the next figure.
\figVehicleConfig

The experiment controlled the initial conditions to reduce the influence of variations in the surrounding traffic environment. The initial conditions were defined by the relative spacing, $ds$, and relative speed, $dv$, of X with respect to A at the activation location. The activation location was at the beginning of the right-turn pavement markings. There X was transitioned from manual operation to tAV mode. The experimental procedure first established the prescribed initial condition and then allowed the automated driving system to independently determine how to execute the mandatory lane change.

For the NC-tAV--tAV Follower experiment, the relative speed between X and A was controlled to remain near zero, whereas relative spacing was varied across three initial-position classes. Vehicle X was positioned near A, near B, or near the midpoint of the A--B gap. Each initial condition was repeated at least five times to identify recurrent behavior across the trials. Although additional repetitions would have increased the available sample, the selected number balanced the need for repeated observations with the operational demands of the labor-intensive field experiment.

To prepare each initial condition, vehicles A, B, and C traveled in the target lane in the downstream-to-upstream order A--B--C, as shown in Figure~\ref{fig:Vehicle Configuration}. They initially traveled at approximately 35~mph before increasing and stabilizing at 40~mph, below the posted speed limit of 45~mph. Vehicles B and C were assigned downstream destinations intended to keep them in the left lane; however, their longitudinal and lane-choice decisions remained under the control of their respective automated driving systems. Thus, the A--B and B--C gaps were not prescribed by the experimental design but developed autonomously through the longitudinal control of tAVs B and C. Accordingly, these gaps represented the relatively small following gaps that the tested tAV system itself maintained under the experimental operating conditions, rather than gaps intentionally created with additional spacing for X to merge. While operating in tAV mode, B and C responded dynamically to the longitudinal adjustments of A and the lane change of X.

Vehicle X traveled in the exclusive right-turn lane and was manually controlled before activation to establish the prescribed spacing and relative-speed condition. Upon reaching the activation location, Auto mode was engaged, and X performed the mandatory lane change under tAV control while the driver continuously oversaw the maneuver. After X merged into the target lane, all four vehicles continued traveling for an additional distance before the trial was concluded.

\subsection{Experimental Sample and Data Coverage}
\label{sec:sample_coverage}

The \textit{tAV dataset} contains 78 controlled mandatory lane-change trials covering three classes of initial spacing conditions: X positioned near A, near the center of the A--B gap, and near B. Each initial condition was repeated multiple times to capture recurrent lane-change behavior under comparable experimental conditions.

The four-vehicle configuration provided four candidate target gaps for X: ahead of A (Gap~0), between A and B (Gap~1), between B and C (Gap~2), and behind C (Gap~3). Table~\ref{tab:target_gap_outcomes} defines these candidate gaps and summarizes the observed target-gap selections. Among the 78 trials, X selected Gap~1 most frequently, with 52 cases. Gap~0 and Gap~2 were selected in 16 and 10 cases, respectively. No trial resulted in rejection of all three available forward gaps or selection of Gap~3 behind C.

\begin{table}[!t]
    \captionsetup{justification=raggedright,singlelinecheck=false}
    \raggedright
    \caption{Candidate target gaps and observed target-gap selections.}
    \label{tab:target_gap_outcomes}
    \begin{tabular}{clc}
        \hline
        \textbf{Target gap} & \textbf{Location relative to the platoon} &
        \textbf{Number of trials} \\
        \hline
        Gap~0 & Ahead of A       & 16 \\
        Gap~1 & Between A and B  & 52 \\
        Gap~2 & Between B and C  & 10 \\
        Gap~3 & Behind C         & 0  \\
        \hline
        \textbf{Total} & & \textbf{78} \\
        \hline
    \end{tabular}
\end{table}
\section{Data and Methods}
\label{sec:data_method}

\subsection{Data Processing}
\label{sec:data_processing}

Each vehicle was instrumented with a single INS unit integrating RTK positioning with GNSS, an accelerometer, and a gyroscope, mounted at the vehicle center. The INS provides time-synchronized position, speed, heading, and acceleration at 20~Hz, with centimeter-level positional accuracy and speed accuracy on the order of 0.01~m/s. Vehicle speed is taken from the INS velocity channel, which is also cross validated by the speed derived from differentiating the processed longitudinal trajectory.

To place all four vehicles in a common longitudinal frame, a reference map of the site was constructed. Vehicle X was driven in tAV mode along each lane in both the NB and SB directions, and its position (i.e., \textit{vehicle center}) was used to approximate the lane centerline. This was based on the consideration that the tAV could presumably localize itself near the lane center more consistently than a human driver, given its advanced sensing and control capabilities. Three repetitions per lane were used to reduce fluctuations and improve accuracy.

The resulting trajectories were densified to 0.05~m resolution by linear interpolation and resampled at 0.1~m intervals to form smooth left- and right-lane centerlines, which were spatially aligned by matching the left-lane point nearest the right-lane start. Lateral lane separation was recorded at every 0.1~m station to capture local road-width variation. Lane widths were \textit{3.7~m and 4.0~m for Site~1 (Northbound) and Site~2 (Southbound), respectively}. The lane boundary was defined as the midpoint between the mapped left- and right-lane centerlines at each longitudinal station.

The raw GNSS positions were converted from geographic coordinates to a local Cartesian frame using geodesic distances under the WGS--84 model. Each vehicle trajectory was then projected onto the left-lane centerline, which served as the common longitudinal reference axis. This produces, for each vehicle and timestamp, a longitudinal coordinate representing position along the road and a lateral coordinate representing offset from the centerline. The trajectories represent the \textit{vehicle center}.

To suppress GNSS noise and referencing artifacts, a 20-point (1~s) moving average is applied to the position data to produce the trajectories for analysis. The same moving average treatment (20-points over a 1~s period) is also applied to the speed data.

\subsection{Key Timestamps}
\label{sec:key_timestamps}

Based on the processed trajectories, each trial was annotated with nine key timestamps that mark the progression of the lane-change maneuver: tAV activation (ACT), lane-change start (LCS), left-edge touching (LET), lane-change crossing (LCC), lane-change end (LCE), and four post-LCE offsets at $+2.5$, $+5$, $+7.5$, and $+10$~s.

ACT, denoted by $t^{*}$, marks the time when X switches from manual mode to tAV mode, also referred to as automation activation. This transition occurs at the automation-activation location. After this time, the tAV takes over immediately while the human driver supervises. ACT was annotated from video records of the vehicle display and defines the controlled initial conditions, $ds$ and $dv$.

LCS marks the onset of the lane-change maneuver. It was identified by tracking the lateral movement of the vehicle, represented by the trajectory of the vehicle center. Specifically, beginning from LCC, the trajectory was searched backward until the smoothed lateral velocity first exceeded a threshold of 0.05~m/s toward the target lane.

LET is the moment when the left edge of the lane-changing tAV first reaches the target-lane boundary, representing the first instant when X physically enters the target lane. At each timestamp, the lateral position of the left-front corner, $Y_{\mathrm{LE}}$, was calculated from the lateral position of the vehicle center, $Y_{\mathrm{C}}$, the vehicle geometry, and the heading angle using Eq.~\ref{eq:left_edge_position}:

\begin{equation}
    Y_{\mathrm{LE}}
    =
    Y_{\mathrm{C}}
    +
    \frac{L}{2}\sin\theta
    +
    \frac{W}{2}\cos\theta ,
    \label{eq:left_edge_position}
\end{equation}

where $L$ and $W$ are the vehicle length and width, respectively, and $\theta$ is the vehicle heading angle relative to the roadway direction. LET was identified as the first timestamp at which the left-front corner reached or crossed the target-lane boundary.

\begin{figure*}[!t]
    \raggedright
    \includegraphics[width=0.72\textwidth]
    {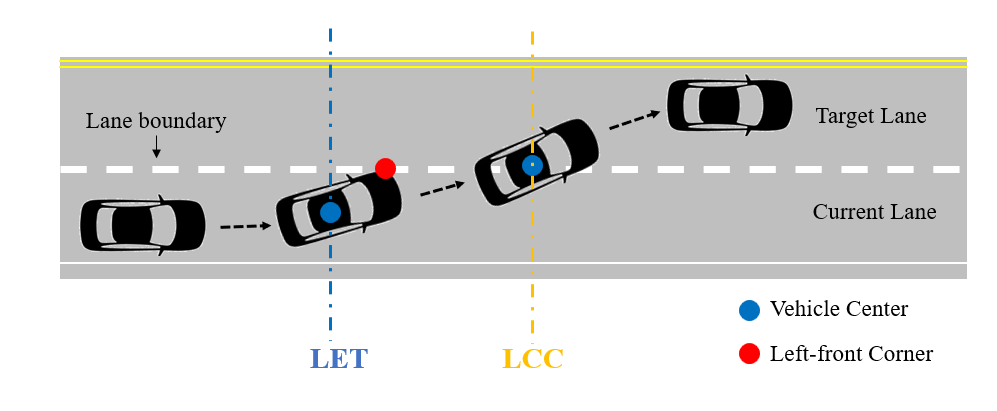}
    \caption{Vehicle position at left-edge touching (LET) and lane-change crossing (LCC). The blue and red points represent the vehicle center and left-front corner, respectively.}
    \label{fig:let_lcc_positions}
\end{figure*}

LCC marks the first timestamp at which the center of X reaches or crosses the target-lane boundary. Figure~\ref{fig:let_lcc_positions} illustrates the distinction between LET and LCC: at LET, the left-front corner of X first reaches the target-lane boundary while the vehicle center remains within the current lane; at LCC, the vehicle center reaches or crosses the boundary.

LCE marks the moment when the lane changer has completely entered the target lane. It was identified as the first timestamp after LCC at which the smoothed lateral velocity fell below 0.05~m/s and remained below this threshold for at least 0.3~s. Four post-LCE timestamps were defined at fixed time offsets of $+2.5$, $+5$, $+7.5$, and $+10$~s from LCE. These timestamps were used to track how the variables of interest evolved after the lane-change maneuver ended. An example of the key timestamps is provided in Fig.~\ref{fig:key-timestamp-example}.

\KeyTimestampFigure

\subsection{Time gap measurement}
\label{sec:time_gap_measurement}

Three time-gap variables are used in the following analyses: the lead gap, the lag gap, and the LC gap. They are all defined based on the time headway between vehicle pairs. The time headway (in the longitudinal direction) between a follower $f$ and a leader $l$ is defined in Eq.~\ref{eq:time_headway}:

\begin{equation}
    h_{fl}(t)
    =
    \frac{x_l(t)-x_f(t)-l_v}{v_f(t)},
    \label{eq:time_headway}
\end{equation}

where $x_l$ and $x_f$ are longitudinal positions (m) represented by the vehicle center (this holds for the remainder of the paper), $l_v$ is vehicle length (m), and $v_f$ is follower speed (m/s). This vehicle-center convention is used throughout the remainder of the paper.

The \textbf{lead gap} ($h_{\mathrm{lead}}$) is the time headway between the lane-changer and its leader in the target lane. We use the term \textit{lead gap} in a loose sense: it measures the time headway between the lane-changer (LCer) and the target-lane leader. For example, although they are in different lanes, when X merges into the gap between vehicles A and B, X may initially be ahead of A and gradually adjust to a position behind A. We nevertheless compute $h_{\mathrm{lead}}$ defined in Eq.~\ref{eq:lead_gap}:

\begin{equation}
    h_{\mathrm{lead}}(t)
    =
    \frac{x_A(t)-x_X(t)-l_v}{v_X(t)},
    \label{eq:lead_gap}
\end{equation}

which is negative when X is ahead of A. Its evolution provides useful information on how the pair relationship develops. Similarly, if X starts behind vehicle B, the lag gap is negative.

The \textbf{lag gap} ($h_{\mathrm{lag}}$) is the time headway between the lane-changer and its follower in the target lane. The \textbf{LC gap} ($h_{\mathrm{LC}}$) is the time headway between the leader and follower in the target lane (pair AB if the tAV merges into Gap~1; pair BC if merging into Gap~2).
\section{Gap Acceptance and Evolution}
\label{sec:gap-evolution}

Lead--lag distribution converges to a region at LC crossing despite different initial conditions.

Despite substantially different initial lead--lag conditions, the three groups of tAV cases evolve toward similar lead--lag states by LC crossing (LCC). Specifically, at LCC, the lead and lag gaps exceed approximately $0.54$~s and $0.65$~s, respectively, and the lag gap is larger than the lead gap in most cases. In addition, the lead--lag distributions become more concentrated at LCC than at activation. Taken together, these results suggest that the state at LCC is not simply a consequence of the initial lead--lag condition. Instead, despite following different adjustment processes, the three groups converge toward a similar lead--lag region. We conjecture that the tAV adjusts its relative position toward a preferred lead--lag operating region by LCC.

\medskip

To examine whether the state at LCC depends on the initial lead--lag condition, we classify the cases into three groups based on the lead and lag gaps at the initial conditions (i.e., at activation, ACT); see Fig.~\ref{fig:lead-lag-evolution} for the plot at ACT. The near-leader group contains cases with $|lead\ gap|<0.5$~s, and the near-follower group contains cases with $|lag\ gap|<0.5$~s. All remaining cases form the near-center group. Although the $0.5$-s threshold is subjective, the grouping is intended to distinguish three qualitatively different initial conditions: a small lead gap, a small lag gap, and a relatively balanced lead--lag split. This grouping allows us to examine whether cases with different initial conditions follow different adjustment processes and whether they evolve toward distinct or similar states by LCC.

After the grouping, we observed the following. For the near-leader group, the lead gap is substantially smaller than the lag gap at activation. Some lead gaps are below zero, indicating that the tAV is actually ahead of the leader and that a substantial adjustment of its relative position is needed. Between activation and LCC, the lead gap increases substantially, with the median increasing from $-0.24$~s to $0.75$~s, whereas the change in the lag gap is much milder. The lead-gap distribution also becomes more concentrated by LCC, as shown in the left plot of Fig.~\ref{fig:gap-distributions}. At LCC, the minimum lead gap is $0.54$~s. Between LCC and LCE, the lead gap continues to increase, but to a much smaller extent. Therefore, the dominant change before LCC occurs in the lead gap, suggesting that the tAV primarily adjusts the lead gap during this stage.

\LeadLagEvolution

The near-follower group exhibits dynamics that mirror those of the near-leader group. At activation, many cases have small lag gaps and large lead gaps. Between activation and LCC, the lag gap increases significantly, with the median increasing from $0.41$~s to $0.85$~s, whereas the LC gap remains relatively stable (the median increasing from $1.47$~s to $1.61$~s; see Fig.~\ref{fig:total-gap-distribution}). Before LCC, most cases satisfy $lag\ gap < lead\ gap$, as shown by the points below the diagonal line in Fig.~\ref{fig:lead-lag-evolution}. However, at LCC, most cases satisfy $lag\ gap \geq lead\ gap$; as indicated by the points on or above the diagonal line. Between LCC and LCE, the lag gap continues to increase relative to the lead gap, resulting in further separation above the diagonal line in Fig.~\ref{fig:lead-lag-evolution} (also reflected by the separation of the lead- and lag-gap distributions in Fig.~\ref{fig:gap-distributions}). At LCC, the minimum lag gap is $0.65$~s. The transition from $lag\ gap < lead\ gap$ to $lag\ gap \geq lead\ gap$ suggests that the tAV tends to maintain a larger lag gap than the lead gap after reaching LCC.

\LeadLagDistribution

The near-center group starts from a more favorable condition, characterized by a more balanced lead--lag split than the near-leader and near-follower groups. At activation, most cases already have lead and lag gaps close to or above the minimum values observed at LCC in the other two groups ($0.54$~s for the minimum lead gap and $0.65$~s for the minimum lag gap). Between activation and LCC, the lag gap increases, but the lead-gap distribution remains largely stable; see Fig.~\ref{fig:gap-distributions}. At LCC, the lag gap exceeds the lead gap in most cases; as shown by the points above the diagonal line in Fig.~\ref{fig:lead-lag-evolution}. Between LCC and LCE, the lag gap continues to increase relative to the lead gap. For this group, because only a modest adjustment is required before LCC, the lead--lag evolution probably reflects the preferred lead--lag distribution of the tAV.

Based on the observations from the three groups, we conjecture that the tAV adjusts toward a preferred lead--lag operating region by LCC. This region is characterized by a lead gap exceeding the minimum lead-gap bound of approximately $0.54$~s and a lag gap exceeding the minimum lag-gap bound of approximately $0.65$~s. Under this conjecture, the near-leader group primarily adjusts the lead gap before LCC because the lead gap is below the minimum lead-gap bound, whereas the near-follower group primarily adjusts the lag gap to meet the minimum lag-gap bound. Interestingly, although the lower bounds are extracted from the near-leader and near-follower groups, they also describe the near-center group: one can see that by LCC, all cases in this group satisfy both bounds. This consistency across the three groups supports the conjecture that the tAV adjusts toward a similar lead--lag operating region despite different initial conditions and adjustment processes. The exact mechanisms governing this behavior are proprietary, and the precise target bounds cannot be verified. Accordingly, we do not attempt to identify their exact values. Instead, we characterize the empirical trends, which provide insight into the impacts of tAV lane-changing behavior on traffic safety and traffic flow, which are to follow later.

\TotalGapDistribution
\section{Risk Assessment}
\label{section:Risk Assessment}
Significant collision risk, predominantly from the target-lane leader, can develop during tAV lane changes, peak at lane entry, and persist beyond their end.

Significant collision risk can exist during the LC process of a tAV. The risk often peaks at LET and decreases thereafter, but it can persist beyond LCE. The dominant risk, when present, originates from front risk arising from the potential conflict between the LCer and the target-lane leader. Specifically, among the 62 cases where tAV merged into Gap 1 or Gap 2, 83.9\% are at risk at LET, about one-third remain at risk at LCE, and 16.1\% remain at risk at LCE+2.5~s. Additionally, front risk dominates in 90.4\% of the at-risk cases at LET and in all remaining at-risk cases at LCE and afterwards. Taken together, these results show that substantial collision risk can develop during the LC process, may remain after the lateral maneuver ends, and is primarily associated with the leader--LCer interaction.

The remark above was obtained through the following process. We first define risk metrics, examine the overall risk evolution and its dominant risk source, and then investigate risk evolution within the three initial-status groups to see how the observed pattern develops.

For risk metrics, we consider a vehicle that merges into a gap between the leader and follower in the target lane. We use \textbf{After Emergency Spacing (AES)} to characterize the projected minimum spacing between two vehicles in the longitudinal direction under an emergency-braking scenario. The AES defined in Eq.~\ref{eq:AES} measures whether the available inter-vehicle spacing at time \(t\) is sufficient to avoid a rear-end collision under a hypothetical emergency-braking scenario. Here, \(x_l\) and \(x_f\) denoted the position of leader and follower vehicle, \(l_v\) is the length of leader vehicle, \(b_0\) is an additional spacing buffer. The scenario is modelled using the following assumptions: (1) At time \(t\), the leading vehicle, travelling at speed \(v_l(t)\), suddenly begins braking with an emergency-deceleration magnitude \(b_l\) and continues braking until it comes to a complete stop. (2) The following vehicle responds after an emergency response time. (3) During this response interval, it continues according to its initial kinematic state, with speed \(v_f(t)\) and acceleration \(a_f(t)\), where \(\lvert a_f(t) \rvert \leq d\). After the response interval, the follower brakes with an emergency-deceleration magnitude \(b_f\). For simplicity, the leading and following vehicles are assumed to have the same emergency-braking capability, such that \( b_l=b_f=d\). This assumption reflects the expectation that the maximum emergency-deceleration capabilities of comparable vehicles do not differ substantially. In this analysis, \(\tau_e = 0.5\ s, d = 0.8\ g, l_v = 5m, b_0 = 2 m\) are considered.

\begin{equation}
AES(t) = (x_l(t)-x_f(t)-l_v-b_0) -(v_f(t)\tau_e +\frac{1}{2}a_f(t)\tau_e^2 +\frac{\left(v_f(t)+a_f(t)\tau_e\right)^2}{2b_f} -\frac{v_l(t)^2}{2b_l}).
\label{eq:AES}
\end{equation}

A negative AES indicates potential collision risk, and a smaller AES represents a less safe condition. For each LC case, $AES_{\mathrm{front}}$ denotes the AES for the leader--LCer pair, whereas $AES_{\mathrm{rear}}$ denotes the AES for the follower--LCer pair. The overall LC safety margin for the LCer is defined as the smaller of the two pairwise safety margins, as given in Eq.~\ref{eq:AES-lc}.

\begin{equation}
    AES_{\mathrm{LC}}
    =
    \min\left(
        AES_{\mathrm{front}},
        AES_{\mathrm{rear}}
    \right).
    \label{eq:AES-lc}
\end{equation}

Thus, $AES_{\mathrm{LC}}$ captures the more critical of the two potential longitudinal conflicts. Accordingly, $AES_{\mathrm{LC}}<0$ indicates potential collision risk, whereas an increase in $AES_{\mathrm{LC}}$ indicates an improvement in the overall safety margin. To make the interpretation more straightforward, front risk and rear risk are defined in Eq.~\ref{eq:pairwise-risk} as

\begin{equation}
    r_{\mathrm{front}}
    =
    -AES_{\mathrm{front}},
    \qquad
    r_{\mathrm{rear}}
    =
    -AES_{\mathrm{rear}}.
    \label{eq:pairwise-risk}
\end{equation}

Under these definitions, positive values indicate potential collision risk, and larger values indicate greater risk. The overall risk of the LCer is therefore defined as the larger of the two pairwise risk values, as given in Eq.~\ref{eq:risk-lc}.

\begin{equation}
    r_{\mathrm{LC}}
    =
    -AES_{\mathrm{LC}}
    =
    \max\left(
        r_{\mathrm{front}},
        r_{\mathrm{rear}}
    \right).
    \label{eq:risk-lc}
\end{equation}

Note that AES captures only the longitudinal interaction between two vehicles. Accordingly, a negative AES becomes physically consequential from LET onward, when the LCer enters the target lane and is in a potential conflict trajectory with its leader or follower. We also report AES before LET to help better understand how the safety margin evolves throughout the LC process.

Based on the risk metrics defined, we first study the 62 cases in which the tAV merged into Gap 1 (52 cases) or Gap 2 (10 cases). The following observations are made.

At LET, 83.9\% of the cases have $AES_{LC}<0$, indicating that the tAV is at risk in most cases at this stage. Between LET and LCE, $AES_{LC}$ increases in almost all cases (61 out of the 62 cases), and its distribution continues to shift upward from LCE through LCE+10~s; see Fig.~\ref{fig:Y1}. Thus, for an LC case, risk peaks at LET and decreases thereafter. However, it does not necessarily disappear when the LC ends. About one-third of the cases remain at risk at LCE, and 16.1\% remain at risk at LCE+2.5~s; see Fig.~\ref{fig:Y2}. A small number of cases remain at risk at later timestamps; see Fig.~\ref{fig:Y2}. The distribution of at-risk period is shown in Fig.~\ref{fig:Y4}. Risk persistence beyond 5~s is common in the 62 cases. In terms of the risk magnitude, a significant fraction (37\%) of the cases has $AES_{LC}<-4$~m (see Fig.~\ref{fig:Y5} for the distribution), suggesting a significant risk magnitude.

\RiskYOneTwo
\RiskYFourFive
%\RiskYOne
%\RiskYTwo
% \RiskYFour
% \RiskYFive

The risk source is predominantly associated with the leader--LCer interaction. Specifically, front risk dominates (i.e., $AES_{\mathrm{front}}<AES_{\mathrm{rear}}$) in 90.4\% of the at-risk cases at LET and in all remaining at-risk cases at LCE and afterwards (see how most points lie above the diagonal line at LET in Fig.~\ref{fig:Y3}).

This front-risk dominance arises through different pathways depending on the initial lead--lag condition. A closer examination by initial-status group reveals how the risk composition evolves differently depending on the initial lead--lag condition.

For the near-leader group, only front risk is present from LET onward, while $AES_{\mathrm{rear}}$ remains positive. Between LET and LCE, $AES_{\mathrm{front}}$ increases (as shown by the rightward shift of the points in the top row of Fig.~\ref{fig:Y3}), indicating that the front risk gradually diminishes but remains the only binding constraint throughout.

By contrast, the near-follower group exhibits a different evolution. Although $AES_{\mathrm{rear}}$ is negative in most cases at LCS, it substantially increases by LET, indicating a substantial improvement in the rear safety margin. However, front risk emerges in all but one case and becomes dominant by LET (see the points on or above the diagonal line at LET in Fig.~\ref{fig:Y3}). By LCC, rear risk has largely disappeared, while front risk remains in most cases. Thus, the dominant risk has shifted from the follower--LCer interaction to the leader--LCer interaction upon lane entry at LET. This shift suggests that the tAV prioritizes reducing rear risk in the early stage of LC process, with rear risk largely eliminated by LCC.

The near-center group exhibits an intermediate pattern. The near-center group begins with a more balanced distribution of $AES_{\mathrm{front}}$ and $AES_{\mathrm{rear}}$ at LCS (cases are distributed on both sides of the diagonal line in the bottom row of Fig.~\ref{fig:Y3}). However, by LET, the distribution resembles that of the near-follower group, with front risk dominating in most cases. The subsequent evolution also follows similar patterns. Therefore, although the near-center group begins from a relatively balanced AES condition, it converges toward the same front-risk-dominant state by LET, as observed in the other two groups.
\RiskYThree
Taken together, the group-specific results reveal a common evolution despite different initial conditions: front risk becomes dominant by LET, and by LCC, rear risk is largely eliminated, while front risk remains dominant through LCE and can persist beyond it.

The common risk pattern can be further understood by relating it to the lead--lag evolution established earlier. The asymmetric lead--lag distribution from LCC onward translates into an asymmetric split between front and rear risk. In particular, the small lead gap at LCC and through LCE translates to substantial risk between the leader and the LCer. Interestingly, some cases retain a noticeable rear safety buffer while bearing front risk (see the points with $AES_{\mathrm{front}}<0$ and $AES_{\mathrm{rear}}>0$ at LCC in Fig.~\ref{fig:Y3}). For example, Case 27 has $AES_{\mathrm{front}}=-4.44$~m and $AES_{\mathrm{rear}}=7.70$~m at LCC. This pattern suggests that the tAV may perceive the front risk as more manageable because its own longitudinal behavior is controllable, whereas the follower's behavior may be more unpredictable. Nevertheless, additional research is needed to determine whether this preference exists in other tAV types.

Other than the 62 cases, we also analyze the 16 cases in which tAV merged into Gap 0. We obtain consistent findings. Specifically, because these cases have no target-lane leader, the relevant intervehicle conflict is only between the LCer and the follower. For most cases, AES is positive at LET, suggesting negligible risk. This is consistent with observations from Gap 1 and Gap 2, where front risk is the binding constraint: with no leader, the primary source of risk is absent.

We note that the reported at-risk percentages pertain only to the controlled experimental scenarios and should not be interpreted as naturalistic estimates of collision-risk frequency. Our purpose is to show that the observed risk features are systematic rather than isolated occurrences. Additionally, AES depends on the assumed reaction time, $\tau$, and braking rate, $d$. Our analysis above uses $\tau=0.5$~s and $d=0.8g$, representing a prompt response and very strong braking. A longer reaction time or weaker braking would produce smaller AES and therefore indicate greater risk. Accordingly, our parameter choice likely underestimates the risk relative to typical real-world conditions. Nevertheless, we believe the key message holds: significant collision risk can develop during tAV lane changes and may persist beyond LC end.

Note that our observations and interpretations apply only to the tested tAV under the tested conditions. Additional tAV systems and scenarios are needed to determine whether the inferred prioritization of rear over front risk generalizes.

\section{Conclusions and Discussions}
\label{sec:conclusion}
Motivated by the lack of systematic empirical studies on the LC behaviors of tAVs, this research makes two major contributions. First, it develops a unique controlled dataset of mandatory lane changes conducted by tAVs, capturing the complete LC process and the interactions between the lane changer and tAV followers. Second, it uses the dataset to extract empirical behavioral and safety insights, focusing on lead–lag gap evolution and collision-risk evolution and source. 
The analyses show that, despite different initial positions, the tested tAVs converge toward a similar lead–lag operating region by LC crossing, with the lag gap generally larger than the lead gap thereafter. Significant projected collision risk can develop during the LC process, typically peaking when the vehicle first enters the target lane, at LET and sometimes persisting beyond LC end. When risk is present, it is predominantly associated with the interaction between the lane changer and the target-lane leader.

The findings have several implications for the modeling and evaluation of tAV lane changes. For example, the LC process should not be represented only by the selected LC gap or a single time stamp in the process, such as LC start or LC crossing. Models should capture the broader process, including the longitudinal adjustment before lane entry, the asymmetric evolution of the lead and lag gaps, and the continued interaction after the lateral maneuver ends. The results also suggest that LC crossing is an important behavioral milestone, by which time the initially different lead–lag conditions have largely converged and rear risk has been substantially reduced.  
The risk findings imply that tAV LC safety should be evaluated over the complete interaction period rather than only at LC start or LC end. In particular, safety metrics should distinguish front and rear risk, identify how the dominant risk source changes during the maneuver, and continue beyond LC end because projected collision risk can persist during post-LC stabilization. The front-risk dominance also suggests that evaluation and modeling should explicitly examine how the tAV manages the tradeoff between creating sufficient space behind and maintaining an adequate safety margin to the target-lane leader.

This research has several  limitations.  The findings should be interpreted within the scope of the experiment. They are based on one tested tAV system, a limited set of initial conditions, and one roadway environment. The specific results therefore should not be generalized directly to broader conditions, other tAV systems, or future versions of the same system, because tAV controllers vary across manufacturers and may evolve through software updates.
Despite these limitations, this research seeks to develop a systematic empirical approach for studying tAV behavior through controlled experiments and thorough vehicle-level analysis. Although the specific behavioral patterns may change as tAV controllers evolve, our results have already revealed tAV behaviors that appear distinct from commonly observed human-driving behavior and may produce significant safety or traffic-flow impacts. These findings therefore identify issues that warrant further attention and the development of appropriate mitigation strategies. The experimental and analytical methodologies are also likely applicable to other tAV systems.
This paper presents an initial set of analysis outcomes from the dataset. Ongoing work examines follower response, LC decision making, and traffic-flow impacts. Future research should test additional tAV manufacturers and software versions and expand the empirical coverage through broader controlled conditions or sufficiently large naturalistic datasets.
Future research is also needed to relate the projected-risk findings to statistical safety outcomes based on observed crash and near-crash records, so that the extent to which the identified AES patterns correspond to actual collision risk can be evaluated.

\section*{AUTHOR CONTRIBUTIONS}
\textbf{Danjue Chen:} Conceptualization, Funding acquisition, Methodology, Investigation, Project administration, Resources, Supervision, Writing--review \& editing.
\textbf{Abhinav Sharma:} Investigation, Data curation, Experimental data collection, Validation, Formal analysis, Visualization, Methodology, Writing--review \& editing.
\textbf{Md Abdullah Al Hasan:} Formal analysis, Visualization, Methodology, Writing--review \& editing.
\textbf{George F. List:} Conceptualization, Supervision, Methodology, Writing--review \& editing.

\section*{DECLARATION OF CONFLICTING INTERESTS}
The authors declared no potential conflicts of interest with respect to the research, authorship, and/or publication of this article.

\section*{FUNDING}
The authors disclosed receipt of the following financial support for the research, authorship, and/or publication of this article: This material is based upon work supported by the National Science Foundation under Award No.~2401555.
\bibliographystyle{trb_chicago}
\bibliography{References}

@article{Wen2023AnalysisData,
    title = {{Analysis of discretionary lane-changing behaviours of autonomous vehicles based on real-world data}},
    year = {2023},
    journal = {Transportmetrica A: Transport Science},
    author = {Wen, Xiao and Huang, Chunxi and Jian, Sisi and He, Dengbo},
    publisher = {Taylor and Francis Ltd.},
    doi = {10.1080/23249935.2023.2288636},
    issn = {23249943}
}

@article{Ali2025AutonomousFollower,
    title = {{Autonomous vehicle lane-changing dynamics and impact on the immediate follower}},
    year = {2025},
    journal = {Analytic Methods in Accident Research},
    author = {Ali, Yasir},
    month = {6},
    volume = {46},
    publisher = {Elsevier Ltd},
    doi = {10.1016/j.amar.2025.100388},
    issn = {22136657}
}

@misc{FordMotorCompany2025BlueCruise,
    title = {{BlueCruise}},
    year = {2025},
    author = {{Ford Motor Company}},
    url = {https://www.ford.com/technology/bluecruise/},
    howpublished = {https://www.ford.com/technology/bluecruise/}
}

@article{Shi2021EmpiricalSettings,
    title = {{Empirical study on car-following characteristics of commercial automated vehicles with different headway settings}},
    year = {2021},
    journal = {Transportation Research Part C: Emerging Technologies},
    author = {Shi, Xiaowei and Li, Xiaopeng},
    month = {7},
    volume = {128},
    publisher = {Elsevier Ltd},
    doi = {10.1016/j.trc.2021.103134},
    issn = {0968090X}
}

@misc{GeneralMotors2025HowCruise,
    title = {{How to Use Super Cruise}},
    year = {2025},
    author = {{General Motors}},
    url = {https://www.chevrolet.com/support/vehicle/driving-safety/driver-assistance/how-to-use-super-cruise},
    howpublished = {https://www.chevrolet.com/support/vehicle/driving-safety/driver-assistance/how-to-use-super-cruise}
}

@article{Sharma2026IntroducingExperimentsb,
    title = {{Introducing the Transitional Autonomous Vehicle Lane-Changing Dataset: Empirical Experiments}},
    year = {2026},
    journal = {arXiv preprint arXiv:2603.05716},
    author = {Sharma, Abhinav and He, Zijun and Chen, Danjue},
    url = {https://doi.org/10.48550/arXiv.2603.05716},
    doi = {10.48550/arXiv.2603.05716},
    arxivId = {2603.05716}
}

@article{Ammourah2024IntroductionExtraction,
    title = {{Introduction to the Third Generation Simulation Dataset: Data Collection and Trajectory Extraction}},
    year = {2024},
    journal = {Transportation Research Record: Journal of the Transportation Research Board},
    author = {Ammourah, Rami and Beigi, Pedram and Fan, Bingyi and Hamdar, Samer H. and Hourdos, John and Hsiao, Chun-Chien and James, Rachel and Khajeh-Hosseini, Mohammdreza and Mahmassani, Hani S. and Monzer, Dana and Radvand, Tina and Talebpour, Alireza and Yousefi, Mahdi and Zhang, Yanlin},
    month = {7},
    doi = {10.1177/03611981241257257},
    issn = {0361-1981}
}

@article{Ali2024InvestigatingDataset,
    title = {{Investigating autonomous vehicle discretionary lane-changing execution behaviour: Similarities, differences, and insights from Waymo dataset}},
    year = {2024},
    journal = {Analytic Methods in Accident Research},
    author = {Ali, Yasir and Sharma, Anshuman and Chen, Danjue},
    month = {6},
    volume = {42},
    publisher = {Elsevier Ltd},
    doi = {10.1016/j.amar.2024.100332},
    issn = {22136657}
}

@article{Zhang2025InvestigationDataset,
    title = {{Investigation of Discretionary Lane-Changing Decisions: Insights From the Third Generation Simulation (TGSIM) Dataset}},
    year = {2025},
    journal = {Transportation Research Record},
    author = {Zhang, Yanlin and Talebpour, Alireza and Mahmassani, Hani S. and Hamdar, Samer H.},
    number = {6},
    month = {6},
    pages = {364--380},
    volume = {2679},
    publisher = {SAGE Publications Ltd},
    doi = {10.1177/03611981251318329},
    issn = {21694052}
}

@misc{TeslaInc.2025ModelSupervised,
    title = {{Model 3 Owner's Manual: Full Self-Driving (Supervised)}},
    year = {2025},
    author = {{Tesla Inc.}},
    url = {https://www.tesla.com/ownersmanual/model3/en_us/GUID-2CB60804-9CEA-4F4B-8B04-09B991368DC5.html},
    howpublished = {https://www.tesla.com/ownersmanual/model3/en{\_}us/GUID-2CB60804-9CEA-4F4B-8B04-09B991368DC5.html}
}

@article{Zheng2014RecentChanging,
    title = {{Recent developments and research needs in modeling lane changing}},
    year = {2014},
    journal = {Transportation Research Part B: Methodological},
    author = {Zheng, Zuduo},
    pages = {16--32},
    volume = {60},
    publisher = {Elsevier Ltd},
    doi = {10.1016/j.trb.2013.11.009},
    issn = {01912615}
}

@article{Mattas2025SafetyExperiments,
    title = {{Safety Perspective on Assisted Lane Changes: Insights from Open-Road, Live-Traffic Experiments}},
    year = {2025},
    author = {Mattas, Konstantinos and Vass, Sandor and Zach{\'{a}}r, Gergely and Ji, Junyi and Gloudemans, Derek and Maggi, Davide and Kriston, Akos and Brahmi, Mohamed and Galassi, Maria Christina and Work, Daniel B and Ciuffo, Biagio},
    doi = {https://doi.org/10.48550/arXiv.2508.09233}
}

@article{Cassidy1999SomeBottlenecks,
    title = {{Some traffic features at freeway bottlenecks}},
    year = {1999},
    author = {Cassidy, Michael J and Bertini, Robert L},
    doi = {10.1016/S0191-2615(98)00023-X}
}

@techreport{NHTSA2022SummarySystems,
    title = {{Summary Report: Standing General Order on Crash Reporting for Level 2 Advanced Driver Assistance Systems}},
    year = {2022},
    author = {{NHTSA}},
    url = {https://www.nhtsa.gov/sites/nhtsa.gov/files/2022-06/ADS-SGO-Report-June-2022.pdf},
    doi = {https://www.nhtsa.gov/sites/nhtsa.gov/files/2022-06/ADS-SGO-Report-June-2022.pdf}
}

@techreport{FHWA2024ThirdFHWA-JPO-24-133,
    title = {{Third Generation Simulation (TGSIM) Data: A Closer Look at The Impacts of Automated Driving Systems on Human Behavior www.its.dot.gov/index.htm Final Report-May 2024 FHWA-JPO-24-133}},
    year = {2024},
    author = {{FHWA}},
    url = {www.its.dot.gov/index.htm}
}

@misc{Waymo2025WaymoDataset,
    title = {{Waymo Open Dataset}},
    year = {2025},
    author = {{Waymo}},
    url = {https://waymo.com/open/}
}
%\bibliography{trb_template}
\end{document}